\documentclass[runningheads]{llncs}

\usepackage{eccv}

\usepackage{eccvabbrv}

\usepackage{graphicx}
\usepackage{booktabs}

\usepackage[accsupp]{axessibility}  

\usepackage{hyperref}

\usepackage{orcidlink}

\usepackage{booktabs}  
\usepackage{tabularx}  
\usepackage{multirow}  
\usepackage{pifont}       
\usepackage{xcolor} 
\usepackage{rotating}
\usepackage[table]{xcolor} 
\usepackage{array}         
\usepackage{bbm}
\usepackage[utf8]{inputenc}

\usepackage{enumitem} 
\usepackage[most]{tcolorbox} 
\tcbuselibrary{skins} 

\usepackage{marvosym} 

\definecolor{bluebg}{RGB}{245, 250, 255}
\definecolor{blueframe}{RGB}{0, 102, 204}

\newcommand{\greencmark}{\scalebox{1.3}{\textcolor{green!70!black}{\ding{51}}}} 
\newcommand{\redxmark}{\scalebox{1.3}{\textcolor{red}{\ding{55}}}}              

\definecolor{highestBase}{HTML}{FFCCCC}  
\definecolor{secondBase}{HTML}{CCE5FF}   
\definecolor{highestDelta}{HTML}{E2F0D9} 

\definecolor{hBase}{HTML}{FFCCCC} 
\definecolor{hTool}{HTML}{CCE5FF} 
\definecolor{hDelta}{HTML}{E2F0D9} 

\definecolor{bgGain}{HTML}{E6FFE6} 
\definecolor{bgLoss}{HTML}{FFE6E6} 

\newcommand{\best}[1]{\textbf{\underline{#1}}}

\newcommand{\benchname}{VTC-Bench\xspace}

\begin{document}

\title{\benchname: Evaluating Agentic Multimodal Models via Compositional Visual Tool Chaining} 

\titlerunning{VTC-Bench: VisualToolChain-Bench}

\author{Xuanyu Zhu\inst{1,7} \and
Yuhao Dong\inst{2}$^\dagger$ \and
Rundong Wang\inst{3} \and Yang Shi\inst{1} \and Zhipeng Wu\inst{4} \and Yinlun Peng\inst{5} \and YiFan Zhang \inst{6}\and Yihang Lou\inst{1} \and Yuanxing Zhang\inst{1} \and Ziwei Liu\inst{2} \and \\ Yan Bai\inst{7}\textsuperscript{,\Letter} \and Yuan Zhou\textsuperscript{\Letter}}

\authorrunning{X. Zhu et al.}


\institute{
$^{1}$PKU \quad
$^{2}$NTU \quad 
$^{3}$USTC\quad
$^{4}$CQU \quad 
$^{5}$NUDT \quad
$^{6}$CASIA \quad
$^{7}$Meituan
}

\maketitle

\renewcommand{\thefootnote}{$\dagger$} 
\footnotetext[1]{Project leader.}      

\renewcommand{\thefootnote}{\Letter}   
\footnotetext[2]{Corresponding authors.} 

\renewcommand{\thefootnote}{\arabic{footnote}} 
\begin{center}
    \large 
    \href{https://github.com/zhuzil/VTC-Bench}{\textcolor{magenta}{\texttt{https://github.com/zhuzil/VTC-Bench}}}
\end{center}
\begin{abstract}
Recent advancements extend Multimodal Large Language Models (MLLMs) beyond standard visual question answering to utilizing external tools for advanced visual tasks. Despite this progress, precisely executing and effectively composing diverse tools for complex tasks remain persistent bottleneck. Constrained by sparse tool-sets and simple tool-use trajectories, existing benchmarks fail to capture complex and diverse tool interactions, falling short in evaluating model performance under practical, real-world conditions. To bridge this gap, we introduce VisualToolChain-Bench~(\benchname), a comprehensive benchmark designed to evaluate tool-use proficiency in MLLMs. To align with realistic computer vision pipelines, our framework features 32 diverse OpenCV-based visual operations. This rich tool-set enables extensive combinations, allowing \benchname to rigorously assess multi-tool composition and long-horizon, multi-step plan execution. For precise evaluation, we provide 680 curated problems structured across a nine-category cognitive hierarchy, each with ground-truth execution trajectories. Extensive experiments on 19 leading MLLMs reveal critical limitations in current models' visual agentic capabilities. Specifically, models struggle to adapt to diverse tool-sets and generalize to unseen operations, with the leading model Gemini-3.0-Pro only achieving 51\% on our benchmark. Furthermore, multi-tool composition remains a persistent challenge. When facing complex tasks, models struggle to formulate efficient execution plans, relying heavily on a narrow, suboptimal subset of familiar functions rather than selecting the optimal tools. By identifying these fundamental challenges, \benchname establishes a rigorous baseline to guide the development of more generalized visual agentic models.

\end{abstract}

\vspace{-0.7cm}

\begin{figure}[t]
    \centering
    \includegraphics[width=0.9\linewidth]{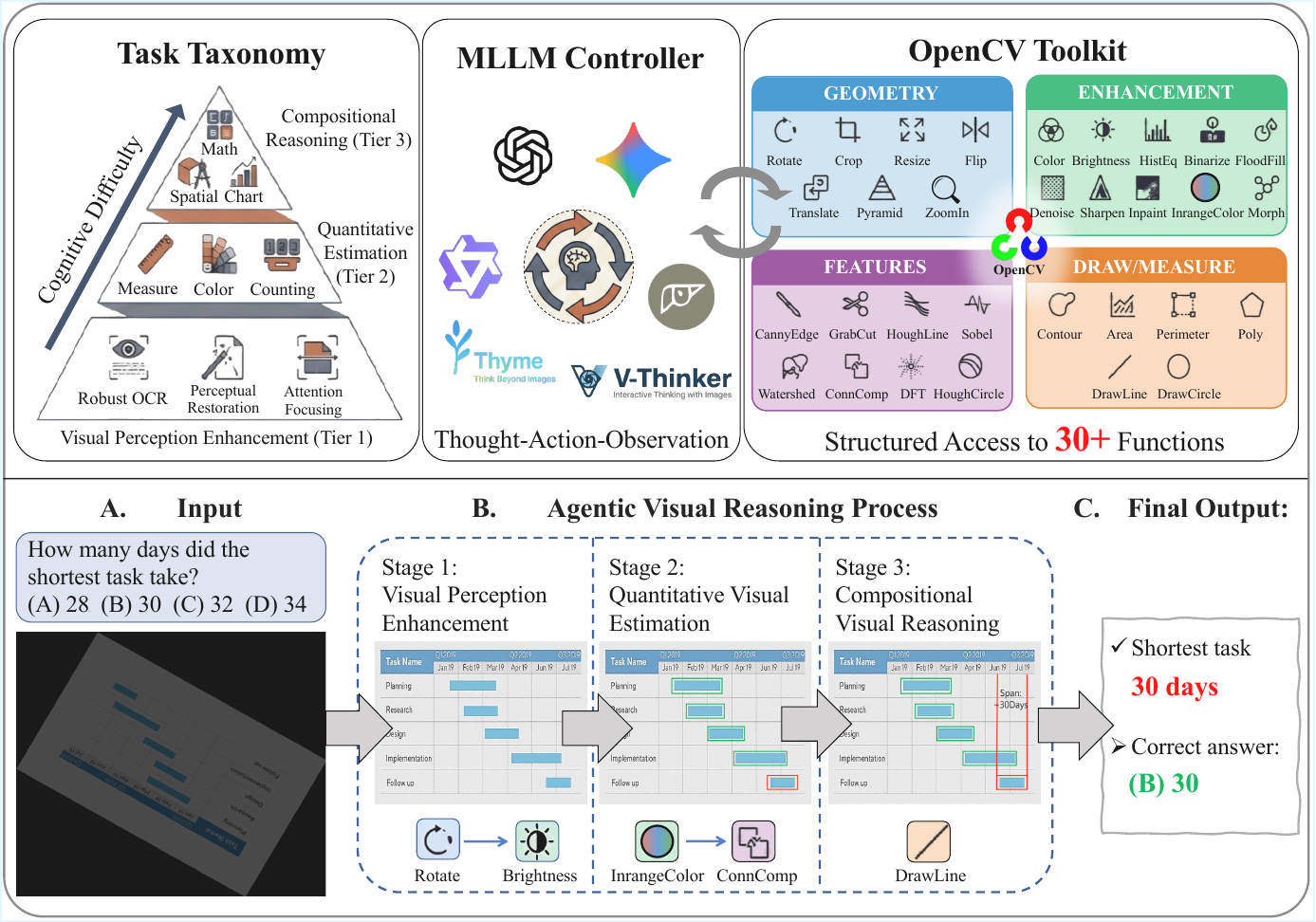}
    \caption{\textbf{Overview of \benchname and the agentic reasoning workflow.} The top panel illustrates the framework architecture, featuring a hierarchical task taxonomy and an MLLM-driven toolkit. The bottom panel exemplifies the multi-stage reasoning trajectory, evolving from visual perception enhancement to compositional reasoning.}
    \label{fig:Teaser}
    \vspace{-0.5cm}
\end{figure}

\section{Introduction}
The rapid evolution~\cite{gemini-2.5-pro,qwen3-vl,google2025gemini3flash,hurst2024gpt4o,openai2025gpt52,kimiteam2026kimik25visualagentic,glm5team2026glm5vibecodingagentic,zhang2025debiasingmultimodallargelanguage,bytedance2026seed2modelcard,google2026gemini,shi2025mavorsmultigranularityvideorepresentation} of Multimodal Large Language Models (MLLMs) has led to remarkable improvements in foundational capabilities such as visual question answering. Building on this progress, recent advancements have expanded their scope by integrating external tools, transforming these models into active, agentic problem solvers. This tool-use ability empowers MLLMs to move beyond basic image understanding to execute complex visual workflows for enhanced image comprehension. By strategically leveraging specialized external visual tools, MLLMs process information more effectively and complete advanced operations. This integration significantly expands their practical skills, making them substantially more versatile for real-world applications.

To assess these emerging agentic capabilities, several benchmarks~\cite{guo2025beyond, li2025tir, su2026agentvista} have been introduced to evaluate how effectively current MLLMs utilize visual tools. However, existing frameworks~\cite{su2026agentvista,guo2025beyond} typically rely on limited tool-sets and simple invocations, rarely testing the complex combinations required for advanced visual reasoning. Furthermore, in practical applications, visual agents must dynamically adapt to a highly diverse array of available tools, and resolving real-world tasks often demands chaining multiple distinct operations together to form a successful execution plan. Failing to capture this necessary diversity and multi-tool composition, current benchmarks obscure the true operational limits of existing models, and this fundamental gap renders them inadequate for guiding the development of more reliable visual agents.

To bridge this critical gap, we introduce VisualToolChain-Bench~(\benchname), a comprehensive benchmark designed to rigorously evaluate the advanced tool-use proficiency of MLLMs on foundational image-based tools. As shown in Fig.~\ref{fig:Teaser}, to emulate authentic computer vision pipelines, our framework integrates 32 distinct visual operations derived from the OpenCV library. These operations serve as the essential building blocks for solving complex visual tasks. By leveraging this versatile tool-set, our benchmark naturally supports advanced tool combinations and multi-step reasoning strategies that reflect real-world challenges. To guarantee a thorough assessment of these capabilities, we constructed 680 meticulously designed problems organized into a nine-level cognitive hierarchy. Furthermore, every problem is paired with a ground-truth execution trajectory to enable the precise evaluation of both intermediate planning and final outcomes. This ensures models are assessed on their underlying logical reasoning rather than merely their final predictions.

We comprehensively evaluate 19 leading MLLMs on \benchname to assess their visual agentic capabilities. Our extensive experiments underscore the highly challenging nature of this benchmark and reveal critical limitations in current models. The overall performance is consistently low, with the top-tier model, like Gemini-3.0-Pro, achieving only 51.2\% on the benchmark. Furthermore, we observe a distinct divergence in tool utilization. While closed-source models demonstrate substantial improvements when equipped with tools, open-source models exhibit minimal gains and sometimes suffer performance degradation. These results highlight a severe pronounced disparity between the current theoretical capability and actual practical proficiency of state-of-the-art models.

To understand the fundamental limitations of current models, we further conduct detailed analysis experiments. Our evaluations reveal that existing models struggle to adapt to diverse tool-sets and generalize to unseen operations. Furthermore, multi-tool composition remains a highly persistent obstacle. We find that models heavily favor a narrow subset of familiar functions instead of actively selecting the optimal tools for a specific task. This strict reliance on known patterns causes significant operational inefficiencies and ultimately leads to execution failures during multi-step reasoning processes. By systematically exposing these specific challenges, \benchname establishes a rigorous baseline to guide the future development of truly generalized visual agents.

\section{Related Work}
\noindent \textbf{Visual Agentic Model}~
Recent Multimodal Large Language Models (MLLMs) are evolving from static textual reasoning toward a dynamic visual agentic paradigm. Early tool-driven approaches~\cite{mmreact, zeng2022socratic} coordinate external vision experts or fixed APIs for basic visual analysis. To enhance perception, interactive attention mechanisms, such as active zooming~\cite{shen2025zoomeye, zhang2025adaptive} and visual masking, are employed to refine inputs. Recent reinforcement learning methods~\cite{zheng2025deepeyes, hong2025deepeyesv2, wang2025pixel, su2025openthinkimg, lai2025mini, wang2025monetreasoninglatentvisual,zhou2025reinforced} optimize these strategies for specific toolset orchestration. However, the reliance on fixed collections of visual parsers fundamentally restricts generalizability. This rigid design confines models to predefined visual scenarios, preventing adaptation to unseen structures.
Programmatic visual manipulation addresses these limitations by utilizing python code as a primitive tool~\cite{gupta2023visual, suris2023vipergpt}. This approach enables on-demand tool construction with complex logic, including loops and conditionals. Advanced frameworks~\cite{hu2024visual, fu2025refocus, vinker2025sketchagent, v-thinker, zhao2025pyvision} dynamically generate code for targeted visual editing, while Thyme~\cite{zhang2025thyme} provides code testing for open-source tool-use models. Leading models, including GPT-o3~\cite{openai2025o3o4mini}, GPT-o4-mini~\cite{openai2025o3o4mini}, and GPT-5.2~\cite{openai2025gpt52}, leverage code execution to construct task-specific tools dynamically. 
Consequently, as state-of-the-art models increasingly embrace this agentic tool-calling paradigm, there is a critical need for a benchmark equipped with a sufficiently diverse tool library to rigorously evaluate their complex, multi-tool compositional capabilities.

\begin{table*}[t]
\centering
\footnotesize
\setlength{\tabcolsep}{10pt} 
\caption{\textbf{Comparison of \benchname with representative multimodal benchmarks.} Compared to existing benchmarks, our work encompasses 32 diverse tools and supports dual interaction paradigms (Code \& Interface). 
\textbf{RT}: Reference tool-call trajectory provided for analysis. 
\textbf{MTC}: Multi-tool composition for a single task. 
\textbf{LHC}: Long-horizon calling with extended execution steps. Tasks necessitate extensive sequential tool invocations. 
\textbf{SFD}: Strict Functional Dependency where prior outputs are mandatory inputs for subsequent tools. As exemplified in Fig.~\ref{fig:Bad_Case_1}, pre-processing serves as a mandatory prerequisite for accurate downstream tool execution.
}
\label{tab:benchmark_comparison}
\resizebox{\linewidth}{!}{
\begin{tabular}{ l c c c c c c c c }
\toprule
\textbf{Benchmark} & \textbf{Paradigm} & \textbf{\#QA} & \textbf{\#Tasks} & \textbf{\#Tools} & \textbf{RT} & \textbf{MTC} & \textbf{LHC} & \textbf{SFD} \\
\midrule
V* \cite{wu2024v} & Static & 191 & 2 & 0 & \redxmark & \redxmark & \redxmark & \redxmark \\
HRBench \cite{wang2025divide} & Static & 200 & 6 & 0 & \redxmark & \redxmark & \redxmark & \redxmark \\
GTA \cite{wang2024gta} & Code & 229 & 3 & 14 & \greencmark & \greencmark & \redxmark & \redxmark \\
TIR-Bench \cite{li2025tir} & Code & 1.2K & 13 & 1/Code & \redxmark & \redxmark & \redxmark & \redxmark \\
Agent-X \cite{ashraf2025agentx} & Code & 828 & 5 & 14 & \greencmark & \greencmark & \redxmark & \redxmark \\
VisualToolBench \cite{guo2025beyond} & Code & 1.2K & 5 & 6 & \greencmark & \greencmark & \redxmark & \redxmark \\
AgentVista \cite{su2026agentvista} & Code & 209 & 7 & 4 & \redxmark & \greencmark & \greencmark & \redxmark \\
\midrule
\rowcolor{gray!15}
\textbf{Ours} & Code \& Interface & 680 & 9 & 32 & \greencmark & \greencmark & \greencmark & \greencmark \\
\bottomrule
\end{tabular}
}
\vspace{-0.4cm}
\end{table*}

\noindent \textbf{Agentic Benchmark in MLLM Benchmark}~
Standard evaluations of multimodal large language models primarily focus on static perception and reasoning. Previous works~\cite{lu2022learn, fu2023mme, lu2023mathvista, shi2025realunifyunifiedmodelstruly,liu2024mmbench, yue2024mmmu,shi2025mmevideoocrevaluatingocrbasedcapabilities,li2025capgeo} test models using static questions and treat vision as a passive input. 
For visual agent evaluation, some studies~\cite{wu2024v, wang2025divide} introduce active visual exploration tasks. However, these early evaluations only examine basic operations like cropping and zooming. Recent research~\cite{wang2024gta, guo2025octopus, li2025tir, guo2025beyond,su2026agentvista,ashraf2025agentx} further advances this field. These methods evaluate multimodal agentic reasoning~\cite{guo2025octopus}, assess image processing capabilities~\cite{li2025tir}, and combine multiple tools for open-ended visual tasks~\cite{guo2025beyond}. 
Despite these advancements, existing benchmarks are inherently constrained by limited tool inventories and lack systematic requirements for compositional multi-tool reasoning, and often fail to capture the nuanced demands of practical, real-world applications.
In contrast, deeply rooted in authentic real-world tasks, our proposed benchmark explicitly targets tool diversity and the complexity of multi-step tool composition. We design 680 problems requiring complex multi-step tool combinations. Models can flexibly call and combine 32 distinct OpenCV~\cite{itseez2014theopencv} tools. 
Agents address these tasks by synthesizing Python code or utilizing our predefined interface,
thereby comprehensively evaluating their ability to generate programmatic solutions for deep visual reasoning. 
A detailed comparison between \benchname and existing benchmarks is presented in Tab.~\ref{tab:benchmark_comparison} to highlight our unique contributions.

\section{VisualToolChain-Bench}
\begin{figure*}[t]
    \centering
    \includegraphics[width=0.9\linewidth]{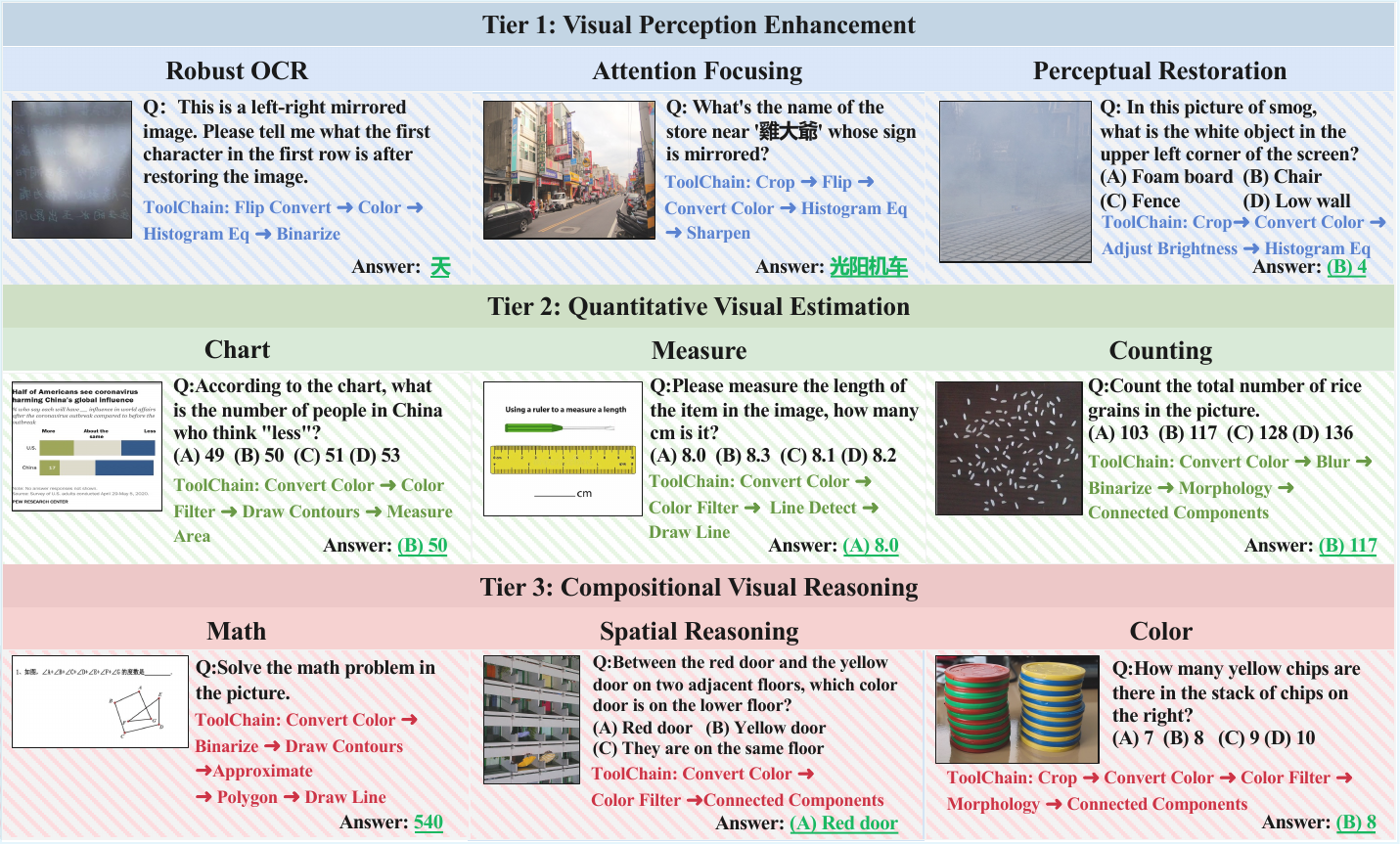}
    \caption{\textbf{Overview of \benchname.} 
    Our benchmark's progressive design structures 9 VQA tasks into a three-tier cognitive hierarchy, evaluating MLLMs' multi-tool orchestration from basic visual recovery to high-level logical deduction.Each task provides a reference toolchain to enable fine-grained diagnostic analysis.}
    \label{fig:overview}
    \vspace{-0.5cm}
\end{figure*}

This section introduces VisualToolChain-Bench(\benchname), with an overview provided in Fig.~\ref{fig:overview}. We first present the benchmark design in Sec.~\ref{sec:design} by establishing a systematic task taxonomy and corresponding toolset. Building upon this foundation, Sec.~\ref{sec:construction} details the benchmark construction process, encompassing data collection, statistical analysis of the dataset, and evaluation metrics.

\subsection{Benchmark Design}\label{sec:design}
\noindent \textbf{Tool Set}~
Due to OpenCV's extensiveness and versatility, we identify OpenCV~\cite{itseez2014theopencv} as our primary tool source to address the sparse tool diversity in the existing benchmarks.
We curated 32 tools, aligning our selection with the standard human cognitive pipeline: initial restoration, feature distillation, and verification. These tools are organized into four functional modules: (1) Geometry for spatial transformations (e.g., rotation and image pyramids); (2) Enhancement for signal optimization (e.g., color space conversion and binarization); (3) Feature Extraction for deriving structural and semantic primitives (e.g., edge detection and watershed segmentation); and (4) Drawing for reasoning verification and attribute quantification (e.g., contour visualization and area measurement). This integrated suite enables controlled visual operations for various MLLMs, with a more comprehensive taxonomy and technical definitions detailed in the App.~\ref{app:tool_set}.

\begin{figure}[t]
  \centering
  \setlength{\tabcolsep}{0pt} 
  
  \begin{minipage}[b]{0.34\linewidth}
    \centering
    \renewcommand{\arraystretch}{1.4}
    \resizebox{\linewidth}{!}{%
      \begin{tabular}[b]{@{}l r@{}}
        \toprule
        \textbf{Statistic} & \textbf{Number} \\
        \midrule
        Total questions & 680 \\
        Multiple-choice & 538 (79.12\%) \\
        Open & 142 (20.88\%) \\
        Questions with image & 680 (100.00\%) \\
        \midrule
        Average chain length & 5.04 \\
        Average unique tools & 4.97 \\
        Total tool calls & 3{,}428 \\
        Min chain length & 1 \\
        Median chain length & 5 \\
        Max chain length & 10 \\
        \midrule
        Average prompt length & 18.52 \\
        \bottomrule
      \end{tabular}%
    }
  \end{minipage}\hfill
  \begin{minipage}[b]{0.55\linewidth}
    \centering
    \includegraphics[width=\linewidth, trim=0.5cm 0.2cm 0.3cm 0.2cm, clip]{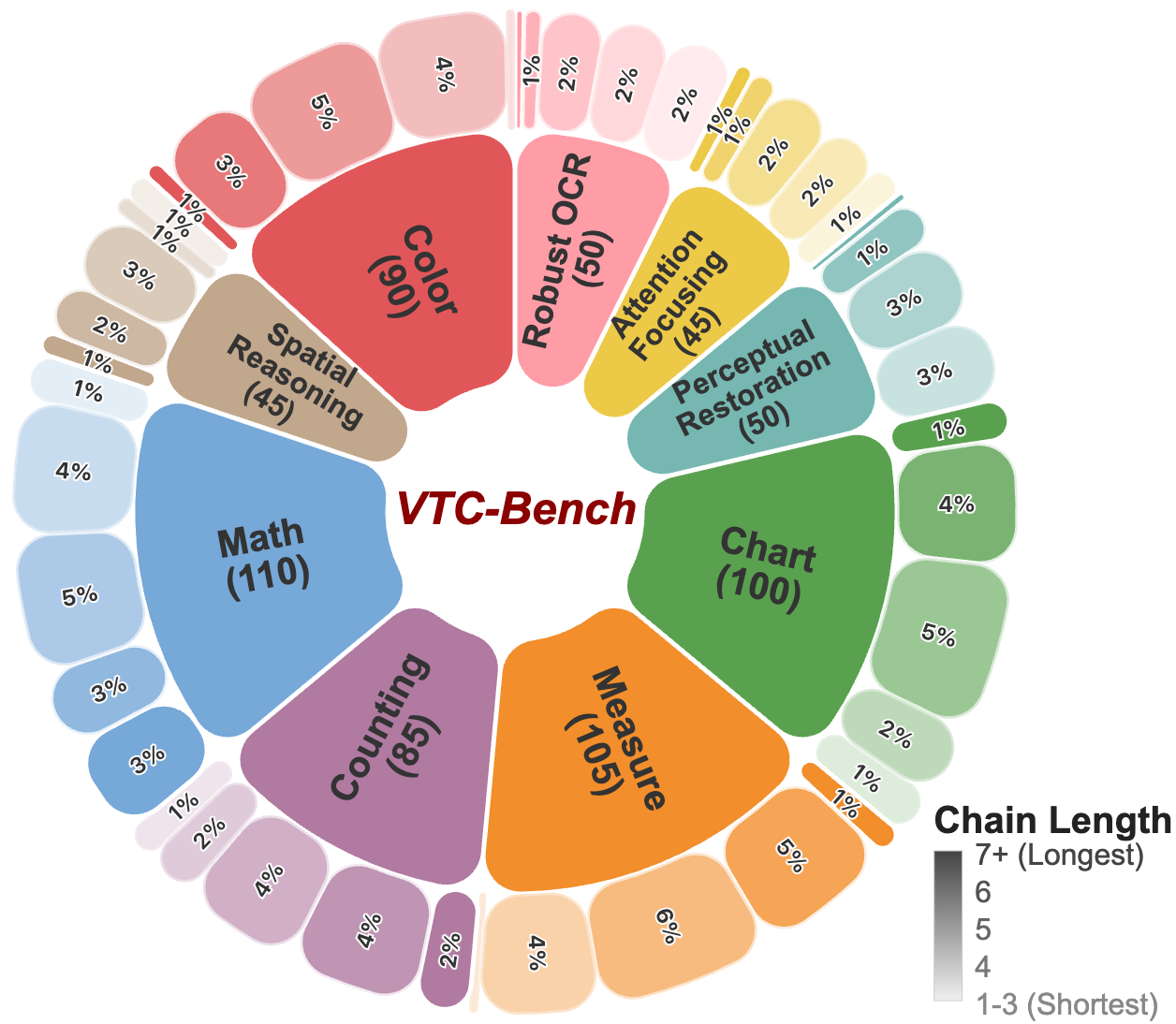}
  \end{minipage}


  \begin{minipage}[t]{0.34\linewidth}
    \setlength{\abovecaptionskip}{0pt} 
    \setlength{\belowcaptionskip}{0pt} 
    \captionof{table}{\strut Statistical summary of \benchname.} 
    \label{tab:benchmark_overview}
  \end{minipage}\hfill
  \begin{minipage}[t]{0.64\linewidth}
    \setlength{\abovecaptionskip}{0pt} 
    \setlength{\belowcaptionskip}{0pt} 
    \captionof{figure}{\strut \textbf{Domain distribution (inner) \& toolchain lengths (outer).} Lengths range from 1-3 to 7+, where darker shades denote longer steps.}
    \label{fig:Category}
  \end{minipage}
  \vspace{-0.5cm}
\end{figure}

\noindent \textbf{Task Design}~
Rather than a fragmented collection of benchmarks, our evaluation suite is structured around a cognitive hierarchy, comprising 9 tasks designed to map the evolution of multimodal agents from passive visual sensing to active constructive reasoning. This hierarchy is organized into three progressive tiers:

    \textit{Tier 1: Visual Perception Enhancement.} This foundational stage comprises \textit{Robust OCR, Perceptual Restoration, and Attention Focusing}. These tasks require models to employ specialized tools to mitigate environmental interference (e.g., haze, low light) and rectify geometric distortions (e.g., rotation). Specifically, \textit{Robust OCR} targets text recognition under synthetic degradation that remains human-readable; \textit{Perceptual Restoration} focuses on scene recovery in adverse conditions such as haze or low light; and \textit{Attention Focusing} emphasizes fine-grained analysis under geometric transformations like rotation or flipping.
    
    \textit{Tier 2: Quantitative Visual Estimation.} Building upon the foundational stage, the tasks of \textit{Measurement, Color, and Counting} evaluate the model's capacity to perceive and precisely quantify physical attributes. Specifically, \textit{Measurement} requires extracting size, position, and shape; \textit{Color} examines the precise extraction of chromatic information; and \textit{Counting} focuses on scene analysis and the strategic invocation of specialized counting tools, rather than relying on the model's intrinsic counting capabilities.
    
    \textit{Tier 3: Compositional Visual Reasoning.} Finally, the \textit{Chart, Math, and Spatial Reasoning} tasks demand complex logical deduction through multi-step tool orchestration. \textit{Chart} is a comprehensive task requiring simultaneous restoration, perception, and inference. \textit{Math} evaluates the construction of auxiliary geometric elements, while \textit{Spatial Reasoning} tests the robust analysis of spatial relations under extreme conditions, such as overexposure or heavy blur.

This hierarchical taxonomy not only ensures comprehensive evaluation dimensions but also reveals the complete cognitive spectrum of multimodal agents, marking a transition from passive visual perception to the sophisticated active constructive capability known as visual agentic model. Consequently, a detailed conceptual overview of the proposed benchmark is illustrated in Fig.~\ref{fig:overview}.

\subsection{Benchmark Construction}\label{sec:construction}
\noindent \textbf{Data collection}~
Our data curation pipeline is guided by several core principles to ensure a rigorous evaluation. We combine web-crawled images with the strategic repurposing of open-source datasets to balance contextual breadth and environmental noise. Rather than relying on original annotations, we synthesize novel instructions from a tool-centric perspective to compel the exploration of latent logical reasoning. This approach transforms static samples into dynamic challenges that require multi-hop execution. Furthermore, we introduce controlled visual perturbations, such as geometric distortions and radiometric noise, to evaluate model robustness. These non-ideal conditions necessitate a transition from passive recognition to active toolchain planning for image restoration.

\noindent \textbf{Verification Protocol}~
\benchname follows a rigorous verification protocol to ensure data integrity. Expert annotators first sanitize images by removing metadata and extracting initial labels. 
We utilize MLLMs including Gemini-3.0-Pro~\cite{google2026gemini} and GPT-5.2~\cite{openai2025gpt52} strictly to validate these manual annotations. Subsequently, Gemini-3.0-Pro drafts the reference toolchains for all samples. Expert researchers then conduct a secondary manual verification on these generated trajectories to determine the finalized ground truth. Finally, a reciprocal auditing phase achieves consensus on accuracy through mutual cross-verification. This rigorous pipeline ultimately yields 680 high-quality robust samples. A technical summary of the entire data collection process is shown in Tab.~\ref{tab:task_overview}.

\begin{table}[t]
    \centering
    \caption{\textbf{Overview of 9 distinct tasks in our benchmark.} All tasks feature human-annotated questions and single-choice answers to ensure rigorous evaluation.}
    \label{tab:task_overview}
    \resizebox{\linewidth}{!}{
    \begin{tabular}{l l c c c}
        \toprule
        \textbf{Task Category} & \textbf{Data Source} & \textbf{QA Construction} & \textbf{Samples} & \textbf{Answer Type} \\
        \midrule
        Attention Focusing      & HRBench~\cite{wang2025divide} \& Web  & Human-annotated & 45 & Open-ended \\
        Chart          & ChartQA~\cite{masry2022chartqa} \& Web  & Human-annotated & 100 & Single-choice \\
        Color          & ColorBench~\cite{liang2025colorbench} \& Web & Human-annotated & 90 & Single-choice \\
        Counting                & Kaggle \& Web                  & Human-annotated & 85 & Single-choice \\
        Math                    & Web \& Manual Construction     &
        Human-annotated & 110 & Single-choice \\
        Measurement             &  Web & Human-annotated & 105 & Single-choice \\
        Perceptual Restoration  & Kaggle  \& Web                       & Human-annotated & 50 & Single-choice \\
        Robust OCR              & OCRBench~\cite{liu2024ocrbench} \& Kaggle & Human-annotated & 50 & Open-ended \\
        Spatial Reasoning       & Kaggle \& Web                  & Human-annotated & 45 & Single-choice \\
        \bottomrule
    \end{tabular}
    }
    \vspace{-0.5cm}
\end{table}

\noindent \textbf{Benchmark Statistics}~\label{sec:benchmark_summary}
As summarized in Tab.~\ref{tab:benchmark_overview}, \benchname\ comprises a diverse set of 680 VQA instances, consisting of 538 multiple-choice and 142 open-ended questions. Each question includes a detailed reference toolchain with an average length of 5.04 steps and 4.97 unique tools, indicating the high complexity of the required operations. Overall, the dataset contains a total of 3,428 tool calls, with chain lengths ranging from 1 to 10 and a median of 5. Furthermore, the average prompt length across all questions is 18.52 words. To better visualize this, Fig.~\ref{fig:Category} illustrates the distribution of each task category, while App.~\ref{app:detaled_task_example} provides representative image and question examples for further clarity.

\subsection{Evaluation Metrics}\label{sec:evaluation_metrics}

We adopt Average Pass Rate (APR) as the primary evaluation metric to represent the proportion of correctly answered questions. 
To analyze tool-use behavior more granularly, we define the \textit{Effective Toolchain} as the minimal sequence of tool calls to produce the final answer. 
This sequence is determined through a backtracking process from the final output to the original input image.

We evaluate the models using three additional metrics: Tool Call Rate (TCR), which measures the proportion of tasks where a model invokes at least one tool; Mean Absolute Error (MAE), which quantifies the discrepancy in length between the predicted and ground-truth toolchains; and Tool Usage Efficiency ($Eff_{\text{tool}}$), which assesses the precision and conciseness of the tool-calling sequences by comparing the number of effective steps to the total predicted steps. 
Mathematically, MAE and $Eff_{\text{tool}}$ are formulated as Eq.~\ref{Eq: Eff_tool}:
\begin{equation}\label{Eq: Eff_tool}
 \text{MAE} = \frac{1}{N} \sum |L_{G, i} - L_{T, i}|, \quad Eff_{\text{tool}} =  \frac{\sum L_{e, i}}{ \sum L_{T, i}}
\end{equation}
where $N$ represents the total number of evaluated samples. The variables $L_{\text{G}, i}$, $L_{\text{T}, i}$, and $L_{\text{e}, i}$ denote the length of the ground-truth toolchain, the total toolchain, and the length of the effective toolchain for the $i$-th sample, respectively.

\begin{table*}[htbp]
  \centering
  \caption{\textbf{Comprehensive evaluation results.} 
  \best{Bold and Underlined} indicates the highest value in each column.
  For Overall and Delta columns: \colorbox{bgGain}{Green} indicates improvement and \colorbox{bgLoss}{Red} indicates degradation compared to base. For subcategories, \colorbox{yellow!40}{Yellow} indicates an improvement of more than 10\% compared to base.}
  \label{tab:comprehensive_results_final}
  \setlength{\tabcolsep}{2.0pt}
  \scriptsize
  \resizebox{\textwidth}{!}{
  \renewcommand{\arraystretch}{1.1}
  \begin{tabular}{lll c c cccccccccc}
    \toprule
    \textbf{Model} & \textbf{Size} &\textbf{Setting} & \textbf{Overall} & \textbf{Delta} & \textbf{OCR} & \textbf{Attn.} & \textbf{Rest.} & \textbf{Chart} & \textbf{Meas.} & \textbf{Count.} & \textbf{Math} & \textbf{Spat.} & \textbf{Color} \\
    \midrule
    \rowcolor{gray!15} \multicolumn{14}{c}{\textbf{Category 1: Proprietary Tool-use Models}} \\
    \midrule
    \textbf{GPT-o3} & --- & Base & 31.91 & & 20.00 & 26.67 & 32.00 & 30.00 & 38.10 & 35.29 & 28.18 & 35.56 & 35.56 \\
     & & Code & \cellcolor{bgGain}36.62 & \cellcolor{bgGain}+4.71 & 28.00 & 31.11 & 34.00 & 40.00 & 38.10 & 34.12 & 30.91 & \cellcolor{yellow!40}51.11 & 42.22 \\
    &  & Inter. & \cellcolor{bgGain}36.76 & \cellcolor{bgGain}+4.85 & 28.00 & 24.44 & 34.00 & 39.00 & 35.24 & 40.00 & 34.55 & 44.44 & 44.44 \\
    \addlinespace
    \textbf{GPT-o4-mini}& --- & Base & 31.18 & & 18.00 & 13.33 & 30.00 & 29.00 & 33.33 & 44.71 & 30.91 & 42.22 & 30.00 \\
     & & Code & \cellcolor{bgGain}33.68 & \cellcolor{bgGain}+2.50 & 12.00 & 11.11 & 30.00 & \cellcolor{yellow!40}41.00 & 31.43 & 35.29 & 40.00 & 46.67 & 37.78 \\
     & & Inter. & \cellcolor{bgGain}32.50 & \cellcolor{bgGain}+1.32 & 24.00 & 13.33 & 28.00 & 30.00 & 33.33 & 40.00 & 33.64 & 46.67 & 35.56 \\
    \addlinespace
    \midrule
    \rowcolor{gray!15} \multicolumn{14}{c}{\textbf{Category 2: Proprietary General-purpose Models}} \\
    \midrule
    \textbf{GPT-4o} & --- & Base & 24.26 & & 18.00 & 15.56 & 22.00 & 25.00 & 25.71 & 25.88 & 22.73 & 26.67 & 30.00 \\
     &  & Code & \cellcolor{bgGain}31.62 & \cellcolor{bgGain}+7.36 & 18.00 & 22.22 & 24.00 & \cellcolor{yellow!40}38.00 & \cellcolor{yellow!40}37.14 & \cellcolor{yellow!40}40.00 & 20.91 & \cellcolor{yellow!40}51.11 & 30.00 \\
     &  & Inter. & \cellcolor{bgGain}33.82 & \cellcolor{bgGain}\best{+9.56} & 20.00 & 22.22 & \cellcolor{yellow!40}36.00 & 31.00 & \cellcolor{yellow!40}41.90 & \cellcolor{yellow!40}40.00 & 24.55 & \cellcolor{yellow!40}48.89 & 37.78 \\
    \addlinespace
    \textbf{Gemini-2.5-Pro} & --- & Base & 38.09 & & 48.00 & 26.67 & 28.00 & 44.00 & 40.95 & 49.41 & 30.00 & 40.00 & 32.22 \\
     & & Code & \cellcolor{bgLoss}36.03 & \cellcolor{bgLoss}-2.06 & 46.00 & \cellcolor{yellow!40}40.00 & 34.00 & 37.00 & 35.24 & 37.65 & 30.00 & 44.44 & 31.11 \\
    &  & Inter. & \cellcolor{bgGain}39.85 & \cellcolor{bgGain}+1.76 & \cellcolor{yellow!40}60.00 & \cellcolor{yellow!40}44.44 & \cellcolor{yellow!40}40.00 & 38.00 & 44.76 & 41.18 & 26.36 & 35.56 & 40.00 \\
    \addlinespace   
    \textbf{GPT-5.2} & --- & Base & 36.03 & & 24.00 & 24.44 & 30.00 & 43.00 & 35.24 & 43.53 & 33.64 & 46.67 & 35.56 \\
    &  & Code & \cellcolor{bgGain}44.56 & \cellcolor{bgGain}+8.53 & \cellcolor{yellow!40}36.00 & 31.11 & 26.00 & \cellcolor{yellow!40}56.00 & \cellcolor{yellow!40}50.48 & 44.71 & \cellcolor{yellow!40}\best{45.45} & 51.11 & 42.22 \\
    &  & Inter. & \cellcolor{bgGain}40.74 & \cellcolor{bgGain}+4.71 & 28.00 & \cellcolor{yellow!40}37.78 & 34.00 & 47.00 & 39.05 & 37.65 & 41.82 & 51.11 & 44.44 \\
    \addlinespace
    \textbf{Gemini-3.0-Flash} & --- & Base & 46.47 & & 44.00 & 51.11 & 46.00 & 57.00 & 50.48 & 52.94 & 27.27 & 60.00 & 40.00 \\
     & & Code & \cellcolor{bgGain}50.59 & \cellcolor{bgGain}+4.12 & \cellcolor{yellow!40}70.00 & 57.78 & 48.00 & 59.00 & 50.48 & 52.94 & 35.45 & 51.11 & 44.44 \\
     & & Inter. & \cellcolor{bgGain}50.74 & \cellcolor{bgGain}+4.27 & \cellcolor{yellow!40}68.00 & 53.33 & \best{50.00} & \best{61.00} & 44.76 & \best{58.82} & \cellcolor{yellow!40}39.09 & 55.56 & 40.00 \\
    \addlinespace
    \textbf{Gemini-3.0-Pro} & --- & Base & 44.41 & & 54.00 & 51.11 & 42.00 & 51.00 & 45.71 & 42.35 & 34.55 & 53.33 & 37.78 \\
     & & Code & \cellcolor{bgGain}\best{51.18} & \cellcolor{bgGain}+6.77 & \cellcolor{yellow!40}\best{74.00} & \cellcolor{yellow!40}62.22 & 38.00 & 60.00 & \best{54.29} & 36.47 & 41.82 & 55.56 & \cellcolor{yellow!40}\best{50.00} \\
    &  & Inter. & \cellcolor{bgGain}51.03 & \cellcolor{bgGain}+6.62 & \cellcolor{yellow!40}70.00 & \cellcolor{yellow!40}\best{73.33} & 38.00 & 59.00 & 50.48 & 35.29 & 37.27 & \cellcolor{yellow!40}\best{71.11} & \cellcolor{yellow!40}\best{50.00} \\
    \addlinespace
    \midrule
    \rowcolor{gray!15} \multicolumn{14}{c}{\textbf{Category 3: Open-source Tool-use Models}} \\
    \midrule
    \textbf{V-Thinker} & 7B & Base & 22.06 & & 8.00 & 11.11 & 22.00 & 27.00 & 23.81 & 27.06 & 18.18 & 28.89 & 24.44 \\
     & & Code & \cellcolor{bgGain}23.82 & \cellcolor{bgGain}+1.76 & 10.00 & 8.89 & 24.00 & 21.00 & 23.81 & \cellcolor{yellow!40}43.53 & 17.27 & \cellcolor{yellow!40}42.22 & 22.22 \\
     & & Inter. & \cellcolor{bgGain}24.41 & \cellcolor{bgGain}+2.35 & 16.00 & 8.89 & \cellcolor{yellow!40}34.00 & 23.00 & 21.90 & 34.12 & 17.27 & \cellcolor{yellow!40}44.44 & 25.56 \\
    \addlinespace
    \textbf{DeepEyes-v2} & 7B & Base & 24.85 & & 22.00 & 11.11 & 18.00 & 27.00 & 38.10 & 21.18 & 20.00 & 28.89 & 26.67 \\
     & & Code & \cellcolor{bgGain}30.00 & \cellcolor{bgGain}+5.15 & 20.00 & 8.89 & 20.00 & 30.00 & 44.76 & \cellcolor{yellow!40}32.94 & 28.18 & \cellcolor{yellow!40}44.44 & 26.67 \\
     & & Inter. & \cellcolor{bgLoss}23.38 & \cellcolor{bgLoss}-1.47 & 18.00 & 6.67 & 28.00 & 16.00 & 28.57 & 30.59 & 18.18 & 35.56 & 27.78 \\
    \addlinespace
    \textbf{DeepEyes} & 7B & Base & 26.47 & & 12.00 & 8.89 & 28.00 & 34.00 & 33.33 & 27.06 & 17.27 & 44.44 & 27.78 \\
     & & Code & \cellcolor{bgGain}27.06 & \cellcolor{bgGain}+0.59 & 22.00 & 8.89 & 28.00 & 30.00 & 32.38 & 36.47 & 17.27 & 40.00 & 25.56 \\
     & & Inter. & \cellcolor{bgGain}29.26 & \cellcolor{bgGain}+2.79 & 22.00 & 6.67 & 28.00 & 29.00 & \cellcolor{yellow!40}43.81 & \cellcolor{yellow!40}37.65 & 11.82 & 46.67 & 33.33 \\
    \addlinespace
    \textbf{Thyme} & 7B & Base & 27.35 & & 8.00 & 8.89 & 34.00 & 29.00 & 41.90 & 35.29 & 18.18 & 35.56 & 24.44 \\
     & & Code & \cellcolor{bgGain}28.82 & \cellcolor{bgGain}+1.47 & 18.00 & 2.22 & 24.00 & 35.00 & 40.00 & 38.82 & 20.91 & 33.33 & 28.89 \\
     & & Inter. & \cellcolor{bgLoss}27.06 & \cellcolor{bgLoss}-0.29 & 14.00 & 4.44 & 24.00 & 33.00 & 33.33 & 34.12 & 20.00 & 40.00 & 28.89 \\
    \addlinespace
    \midrule
    \rowcolor{gray!15}\multicolumn{14}{c}{\textbf{Category 4: Open-source General-purpose Models}} \\
    \midrule
    \textbf{Qwen3-VL-Instruct} & 8B & Base & 30.74 & & 32.00 & 11.11 & 28.00 & 27.00 & 31.43 & 38.82 & 31.82 & 35.56 & 33.33 \\
     & & Code & \cellcolor{bgLoss}28.24 & \cellcolor{bgLoss}-2.50 & 26.00 & 17.78 & 20.00 & 27.00 & 39.05 & 29.41 & 23.64 & 17.78 & 37.78 \\
     & & Inter. & \cellcolor{bgLoss}28.68 & \cellcolor{bgLoss}-2.06 & 30.00 & 17.78 & 34.00 & 20.00 & 30.48 & 42.35 & 24.55 & 26.67 & 31.11 \\
    \addlinespace
    \textbf{Qwen3-VL-Thinking} & 8B & Base & 30.29 & & 20.00 & 11.11 & 24.00 & 33.00 & 40.00 & 35.29 & 29.09 & 35.56 & 28.89 \\
    &  & Code & \cellcolor{bgLoss}26.32 & \cellcolor{bgLoss}-3.97 & 12.00 & 6.67 & 22.00 & 29.00 & 32.38 & 36.47 & 22.73 & 20.00 & 34.44 \\
    &  & Inter. & \cellcolor{bgLoss}29.41 & \cellcolor{bgLoss}-0.88 & 20.00 & 15.56 & 22.00 & 34.00 & 32.38 & 37.65 & 24.55 & 28.89 & 35.56 \\
    \addlinespace
    \textbf{Qwen3-VL-Instruct} & 32B & Base & 32.79 & & 28.00 & 8.89 & 26.00 & 37.00 & 31.43 & 48.24 & 27.27 & 44.44 & 34.44 \\
     & & Code & \cellcolor{bgLoss}30.59 & \cellcolor{bgLoss}-2.20 & 28.00 & 15.56 & 20.00 & 32.00 & 31.43 & 34.12 & 33.64 & 31.11 & 35.56 \\
     & & Inter. & \cellcolor{bgGain}33.09 & \cellcolor{bgGain}+0.30 & 34.00 & 13.33 & 16.00 & 21.00 & 37.14 & 51.76 & 29.09 & 51.11 & 38.89 \\
    \addlinespace
    \textbf{Qwen3-VL-Thinking} & 32B & Base & 32.94 & & 20.00 & 4.44 & 28.00 & 32.00 & 40.95 & 43.53 & 34.55 & 42.22 & 32.22 \\
     & & Code & \cellcolor{bgGain}33.09 & \cellcolor{bgGain}+0.15 & 22.00 & 8.89 & 26.00 & 41.00 & 35.24 & 43.53 & 31.82 & 33.33 & 35.56 \\
     & & Inter. & \cellcolor{bgGain}33.97 & \cellcolor{bgGain}+1.03 & 30.00 & \cellcolor{yellow!40}17.78 & 28.00 & 32.00 & 35.24 & 44.71 & 31.82 & 40.00 & 37.78 \\
    \addlinespace
    \textbf{Qwen3-VL-Instruct} & 30B-A3B & Base & 33.09 & & 42.00 & 15.56 & 22.00 & 23.00 & 41.90 & 41.18 & 32.73 & 42.22 & 32.22 \\
     & & Code & \cellcolor{bgLoss}31.03 & \cellcolor{bgLoss}-2.06 & 42.00 & 22.22 & 32.00 & 31.00 & 31.43 & 36.47 & 23.64 & 37.78 & 28.89 \\
     & & Inter. & \cellcolor{bgLoss}31.91 & \cellcolor{bgLoss}-1.18 & 40.00 & 24.44 & 28.00 & 25.00 & 37.14 & 40.00 & 28.18 & 44.44 & 25.56 \\
    \addlinespace
    \textbf{Qwen3-VL-Thinking} & 30B-A3B & Base & 29.56 & & 16.00 & 8.89 & 22.00 & 31.00 & 30.48 & 44.71 & 24.55 & 46.67 & 32.22 \\
    &  & Code & \cellcolor{bgLoss}28.24 & \cellcolor{bgLoss}-1.32 & 18.00 & 13.33 & 22.00 & 35.00 & 36.19 & 23.53 & 27.27 & 40.00 & 27.78 \\
    &  & Inter. & \cellcolor{bgGain}33.24 & \cellcolor{bgGain}+3.68 & \cellcolor{yellow!40}28.00 & 15.56 & 24.00 & 29.00 & 39.05 & 43.53 & 30.00 & 48.89 & \cellcolor{yellow!40}34.44 \\
    \addlinespace
    \textbf{Qwen3-VL-Instruct} & 235B & Base & 36.32 & & 42.00 & 11.11 & 28.00 & 33.00 & 40.95 & 48.24 & 35.45 & 35.56 & 38.89 \\
    &  & Code & \cellcolor{bgLoss}33.68 & \cellcolor{bgLoss}-2.64 & 36.00 & 15.56 & 28.00 & 26.00 & 40.00 & 42.35 & 30.00 & \cellcolor{yellow!40}46.67 & 35.56 \\
    &  & Inter. & \cellcolor{bgLoss}34.85 & \cellcolor{bgLoss}-1.47 & 40.00 & \cellcolor{yellow!40}22.22 & 20.00 & 32.00 & 38.10 & 48.24 & 27.27 & 37.78 & 41.11 \\
    \addlinespace
    \textbf{Qwen3-VL-Thinking} & 235B & Base & 35.15 & & 34.00 & 8.89 & 24.00 & 38.00 & 39.05 & 45.88 & 32.73 & 37.78 & 38.89 \\
    &  & Code & \cellcolor{bgLoss}34.41 & \cellcolor{bgLoss}-0.74 & 36.00 & \cellcolor{yellow!40}22.22 & 26.00 & 37.00 & 36.19 & 35.29 & 34.55 & 37.78 & 36.67 \\
    &  & Inter. & \cellcolor{bgGain}38.09 & \cellcolor{bgGain}+2.94 & 40.00 & \cellcolor{yellow!40}24.44 & \cellcolor{yellow!40}38.00 & 35.00 & 32.38 & 52.94 & 34.55 & \cellcolor{yellow!40}48.89 & 38.89 \\
    \addlinespace
    \bottomrule
  \end{tabular}
  }
\end{table*}

\section{Experiment}
\begin{figure}[t]
    \centering
    \includegraphics[width=0.95\linewidth]{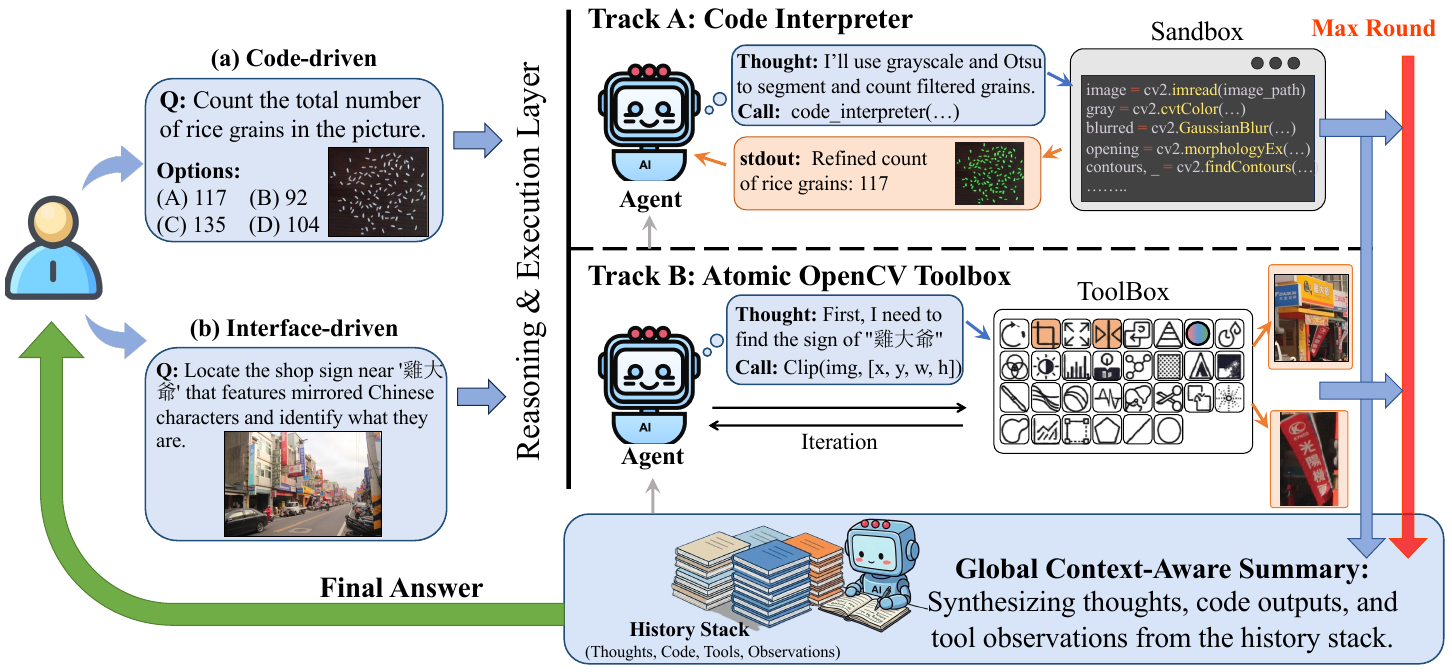}
    \caption{\textbf{Overview of workflows for two tool-use protocols of LLM agents during evaluation.} The agent iteratively interacts with tools within the Max Round constraint, ultimately synthesizing tool execution results to generate the final answer.}
    \label{fig:WorkFlow}
    \vspace{-0.3cm}
\end{figure}

\subsection{Experiment Setup}\label{sec:experiment_setup}
\noindent \textbf{Model Selection}~
We evaluate 19 mainstream MLLMs, encompassing both proprietary and open-source models with diverse tool-use training backgrounds. We stratified these evaluated models into four distinct categories. The first category comprises proprietary tool-use models, featuring closed-source systems with robust native capabilities like GPT-o3~\cite{openai2025o3o4mini} and GPT-o4-mini~\cite{openai2025o3o4mini}. The second category incorporates leading proprietary general-purpose models, such as GPT-5.2~\cite{openai2025gpt52}, Gemini-3.0-Pro~\cite{google2026gemini}, Gemini-3.0-Flash~\cite{google2025gemini3flash}, Gemini-2.5-Pro~\cite{gemini-2.5-pro}, and GPT-4o~\cite{openai2025o3o4mini}. Transitioning to the open-source domain, the third group consists of dedicated open-source tool-use models, including specialized architectures such as DeepEyes~\cite{zheng2025deepeyes}, Thyme~\cite{zhang2025thyme}, DeepEyesV2~\cite{hong2025deepeyesv2}, and V-Thinker~\cite{v-thinker}. Finally, for open-source general-purpose models, we evaluate both the Instruct and Thinking variants of Qwen3-VL~\cite{qwen3-vl} across parameter scales ranging from 8B to 235B.

\noindent \textbf{Evaluation Framework}~
To measure the benefits of tool invocation, we compare direct answering baselines against tool-augmented reasoning across both interface-driven and code-driven paradigms.
This comparison reveals how a model's programming proficiency influences overall orchestration success. 
The evaluation workflow is illustrated in Fig.~\ref{fig:WorkFlow}.
We utilize the Qwen-Agent~\cite{qwen_agent} framework to manage both code and interface implementations for models with robust native tool-calling support. 
Conversely, we employ the Thyme~\cite{zhang2025thyme} framework for models with limited native tool-calling proficiency, applying this approach primarily to open-source tool-use models to facilitate code generation. 
To facilitate this process, we provide detailed tool descriptions that explicitly outline usage instructions and parameter specifications, ensuring that even models lacking specialized OpenCV training can comprehend and invoke the available interfaces. 
Finally, we implement a comprehensive dual-stream evaluation protocol, integrating deterministic rule-based matching with a GPT-4o~\cite{hurst2024gpt4o} LLM-as-a-Judge paradigm. 
Detailed prompt templates are provided in the App.~\ref{app:prompts}.

\subsection{Main Result}\label{sec:result}

\noindent \textbf{The benchmark presents a significant challenge to current MLLMs.} Performance in the base setting remains relatively low, with scores ranging from 22.06\% to 46.47\%. Gemini-3.0-Flash~\cite{google2025gemini3flash} achieves the highest base score of 46.47\%.
Most models cluster around the 30\% mark.
Notably, simply scaling up model parameters does not inherently solve this critical bottleneck; for instance, Qwen3-VL-235B-A22B~\cite{qwen3-vl} achieves a performance improvement from 30.74 to 36.32\% compared to Qwen3-VL-8B; however, it still struggles to surpass 40\% even with tool augmentation.
This universal performance ceiling underscores that transitioning from passive visual perception to active, multi-step tool orchestration remains an unsolved frontier, even for advanced industry-leading architectures.

\noindent \textbf{Proprietary models significantly outperform open-source models.}
As illustrated in Tab.~\ref{tab:comprehensive_results_final}, our evaluation reveals a pronounced performance disparity between proprietary and open-source models, particularly concerning their capacity to leverage external tools. 
Proprietary models not only establish robust baselines but also exhibit substantial performance surges when augmented with tool-use capabilities; 
notably, GPT-4o achieves a remarkable gain of +9.56\% under the interface setting, while GPT-5.2 realizes an +8.53\% improvement in the code setting. Furthermore, Gemini-3.0-Pro attains the highest overall score of 51.18\% with tools, underscoring the advanced reasoning and execution stability inherent in closed-source paradigms. 
Conversely, open-source models frequently fail to effectively harness tool integration, often experiencing limited gains when augmented with tool-use capabilities, indicating a considerable gap between open-source and proprietary models in native tool-use capabilities.

\noindent \textbf{Models perform better at coarse-grained tool invocation, and intrinsic perception is the prerequisite for tool use.}
In visual perception enhancement tasks, models exhibit strong performance, with GPT-4o boosting \textit{Perceptual Restoration} by +14.00\% via interface.
These tasks require only basic, \textit{coarse-grained} tool operations, such as rotation or mirroring, to achieve significant improvements.
Conversely, quantitative visual estimation yields bifurcated outcomes, with noticeable degradation in high-precision perception tasks. Because Tier 2 rigorously tests precise tool selection and parameter configuration, it exposes models' inability to execute \textit{fine-grained} tool manipulations.
Finally, in compositional visual reasoning tasks, models achieve remarkable leaps, highlighted by a +17.78\% improvement in Gemini-3.0-Pro's \textit{Spatial Reasoning} score.
This resurgence stems from the comprehensive nature of Tier 3, which simultaneously challenges intrinsic perception and tool proficiency.
Notably, models with stronger foundational perception excel here, underscoring that robust intrinsic perception is an indispensable prerequisite for advanced tool calling.

\noindent \noindent \textbf{The proposed toolset provides robust support for complex visual reasoning.} 
Notably, contrary to the common assumption that code execution is inherently superior, our findings demonstrate that interface-based invocation is comparable to code-based approaches overall. 
Specifically, among the five proprietary general-purpose models evaluated, three exhibited better performance with interface-based methods. 
These quantitative results further validate that our structured 32-tool library effectively meets the requirements of visual agents. 

\subsection{Analysis Experiments}\label{sec:tool_use_analysis}
\begin{figure}[t]
    \centering
    \includegraphics[width=0.95\linewidth]{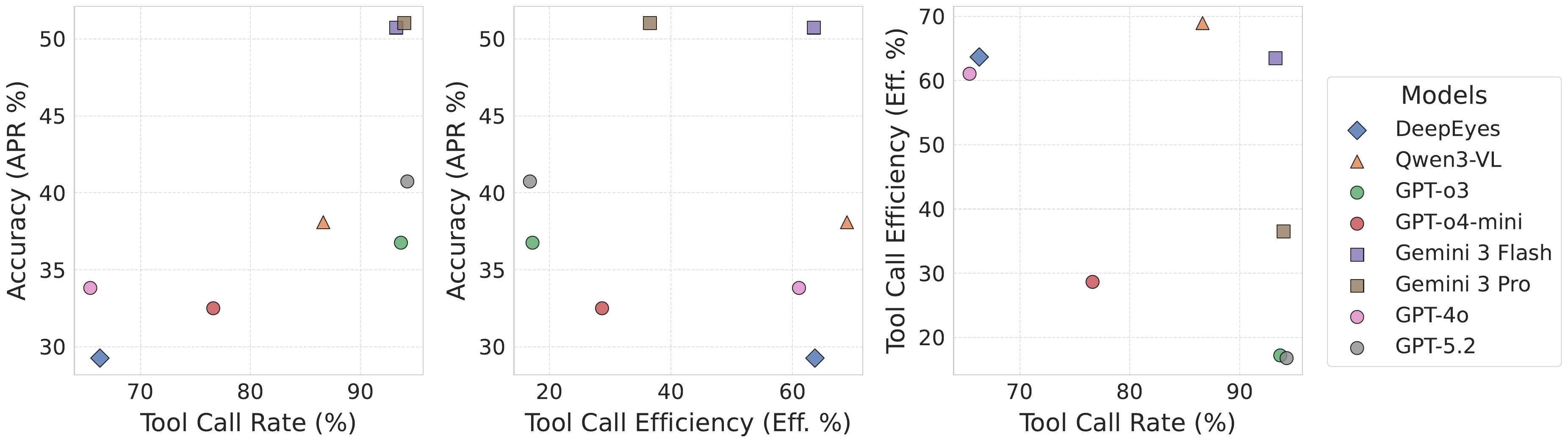}
    \caption{\textbf{Relationships among Tool Call Rate, Tool Call Efficiency, and APR.} The panels illustrate APR vs. Tool Call Rate (left), APR vs. Tool Call Efficiency (middle), and Tool Call Rate vs. Tool Call Efficiency (right).}
    \label{fig:tool_performance_analysis}
    \vspace{-0.5cm}
\end{figure}

\noindent \textbf{Analysis of Tool Utilization and Efficiency.} The left panel of Fig.~\ref{fig:tool_performance_analysis} reveals a positive correlation between Tool Call Rate~(TCR) and Average Pass Rate~(APR), validating the utility of active tool invocation. Conversely, the efficiency analysis in the right panel indicates that general-purpose models achieve superior efficiency even with lower TCR. Specialized models cluster in the high-redundancy region at the bottom right. Their specialized training incentivizes frequent tool usage but often causes them to neglect their inherent perceptual capabilities. Consequently, a disconnect exists between perception and tool calling in these architectures. Our results indicate that current specialized models fail to integrate these two abilities effectively.


\begin{table}[t]
\centering
\caption{\textbf{Statistical analysis of tool invocation performance.} Beyond the Tool-calling Rate(TCR) and Average Pass Rate (APR), we provide a fine-grained evaluation categorized by total attempted tool calls (All Calls) and the effective toolchain (Effective Calls). For both categories, we detail the average number of calls (Avg. T.), the number of unique tools used (Uniq. T.), and the Mean Absolute Error (MAE) from the ground-truth length. Finally, the Tool Usage Efficiency  ($Eff_{\text{tool}}$, denoted as Eff.) represents the ratio of effective steps to total attempted steps.}
\label{tab:tool_analysis}
\resizebox{\columnwidth}{!}{%
\setlength{\tabcolsep}{3pt} 
\begin{tabular}{l cc ccc ccc c}
\toprule
\multirow{2}{*}{\textbf{Model}} & \multirow{2}{*}{\textbf{TCR (\%)}} & \multirow{2}{*}{\textbf{APR (\%)}} & \multicolumn{3}{c}{\textbf{All Calls}} & \multicolumn{3}{c}{\textbf{Effective Calls}} & \multirow{2}{*}{\textbf{Eff. (\%)}}\\
\cmidrule(lr){4-6} \cmidrule(lr){7-9}
& & & Avg. T. & Uniq. T. & MAE & Avg. T. & Uniq. T. & MAE & \\
\midrule
DeepEyes         & 66.32 & 29.26 & 1.90  & 1.04 & 4.09 & 1.21 & 0.90 & 3.97 & 63.68 \\
Qwen3-VL-235B    & 86.62 & 38.09 & 2.19  & 1.77 & 3.02 & 1.51 & 1.47 & 3.36 & 68.95 \\
GPT-5.2          & 94.26 & 40.74 & 13.23 & 4.08 & 9.96 & 2.22 & 1.97 & 3.01 & 16.78 \\
GPT-o3           & 93.68 & 36.76 & 9.77  & 3.47 & 6.98 & 1.68 & 1.51 & 3.32 & 17.20 \\
GPT-o4-mini      & 76.62 & 32.50 & 3.63  & 1.77 & 3.63 & 1.04 & 1.02 & 3.76 & 28.65 \\
Gemini-3.0-Pro   & 93.97 & 51.03 & 5.04  & 2.72 & 3.87 & 1.84 & 1.76 & 3.19 & 36.51 \\
GPT-4o           & 65.44 & 33.82 & 2.26  & 1.81 & 3.91 & 1.38 & 1.26 & 3.82 & 61.06 \\
Gemini-3.0-Flash & 93.24 & 50.74 & 1.89  & 1.38 & 3.26 & 1.20 & 1.19 & 3.64 & 63.49 \\
\bottomrule
\end{tabular}
}
\end{table}

\begin{figure}[t]
    \centering
    \includegraphics[width=0.95\linewidth]{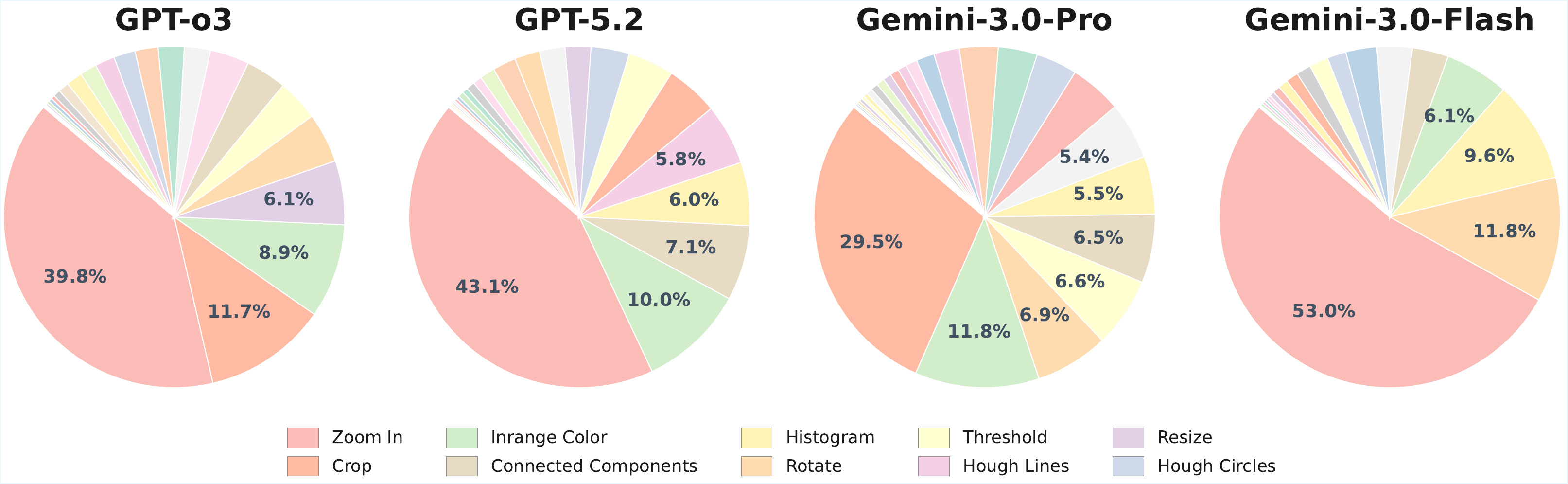}
    \caption{\textbf{Distribution of tool utilization across different models.} The usage patterns remain largely consistent across models, with the most frequently invoked tools including \textit{Zoom In, Inrange Color, Rotate, Histogram, and Connected Components}.}
    \label{fig:tool_call_pie}
\end{figure}

\noindent \textbf{Uneven Distribution of Tool Utilization.} Furthermore, we analyze the tool-use distribution within representative models from the Gemini and GPT series. Fig.~\ref{fig:tool_call_pie} reveals that a small number of tools account for a disproportionately large percentage of total calls. For instance, models frequently utilize basic tools such as crop, zoom in, and rotate. This concentration suggests that existing models possess limited diversity in their tool selection. They primarily rely on simple and common operations that were likely heavily emphasized during their training. These empirical results indicate that most current models still lack the capability for diverse and complex tool invocation.

\begin{figure}[t]
    \centering
    \begin{subfigure}{0.48\textwidth}
        \centering
        \includegraphics[width=\linewidth]{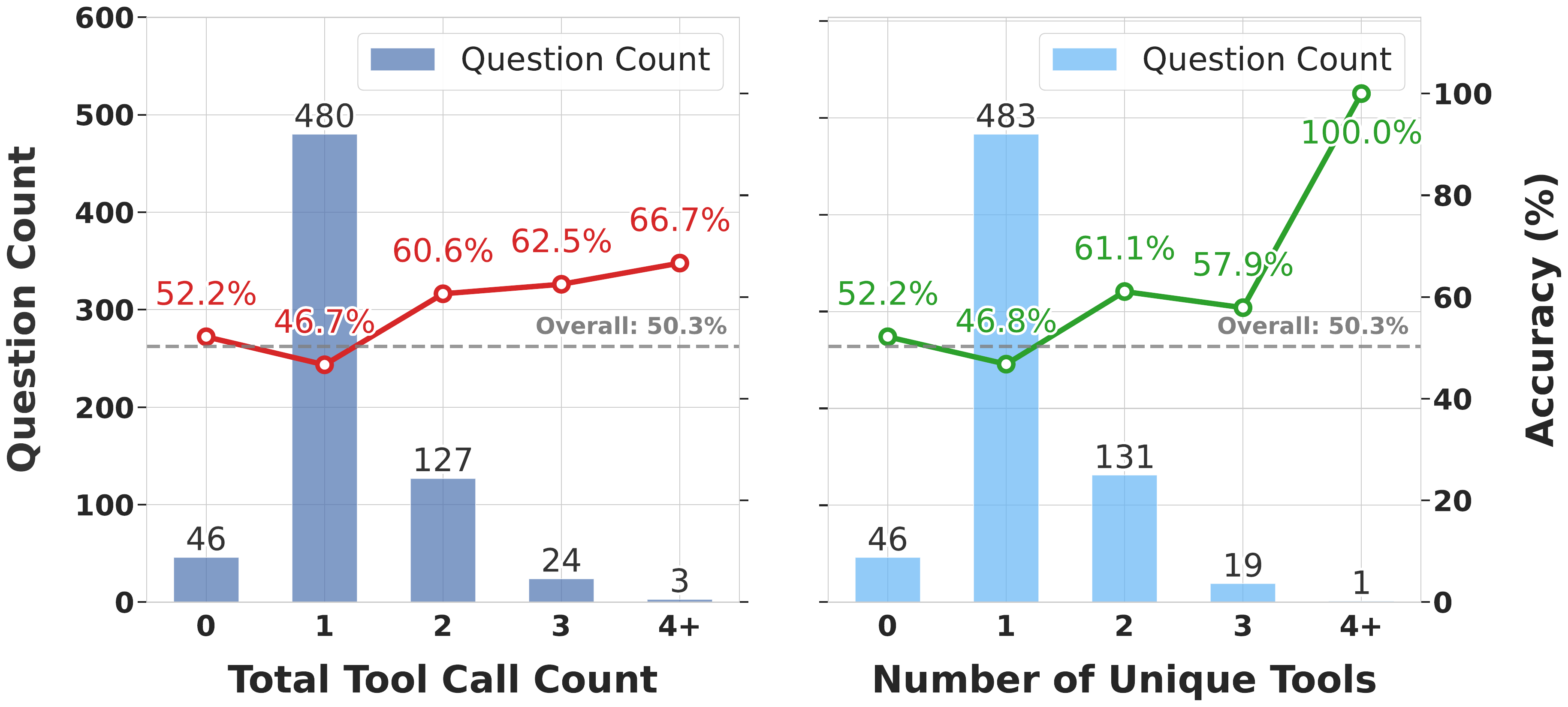}
        \caption{\textbf{Sample distribution vs. Accuracy.}}
        \label{fig:tool_analysis}
    \end{subfigure}
    \hfill 
    \begin{subfigure}{0.48\textwidth}
        \centering
        \includegraphics[width=\linewidth]{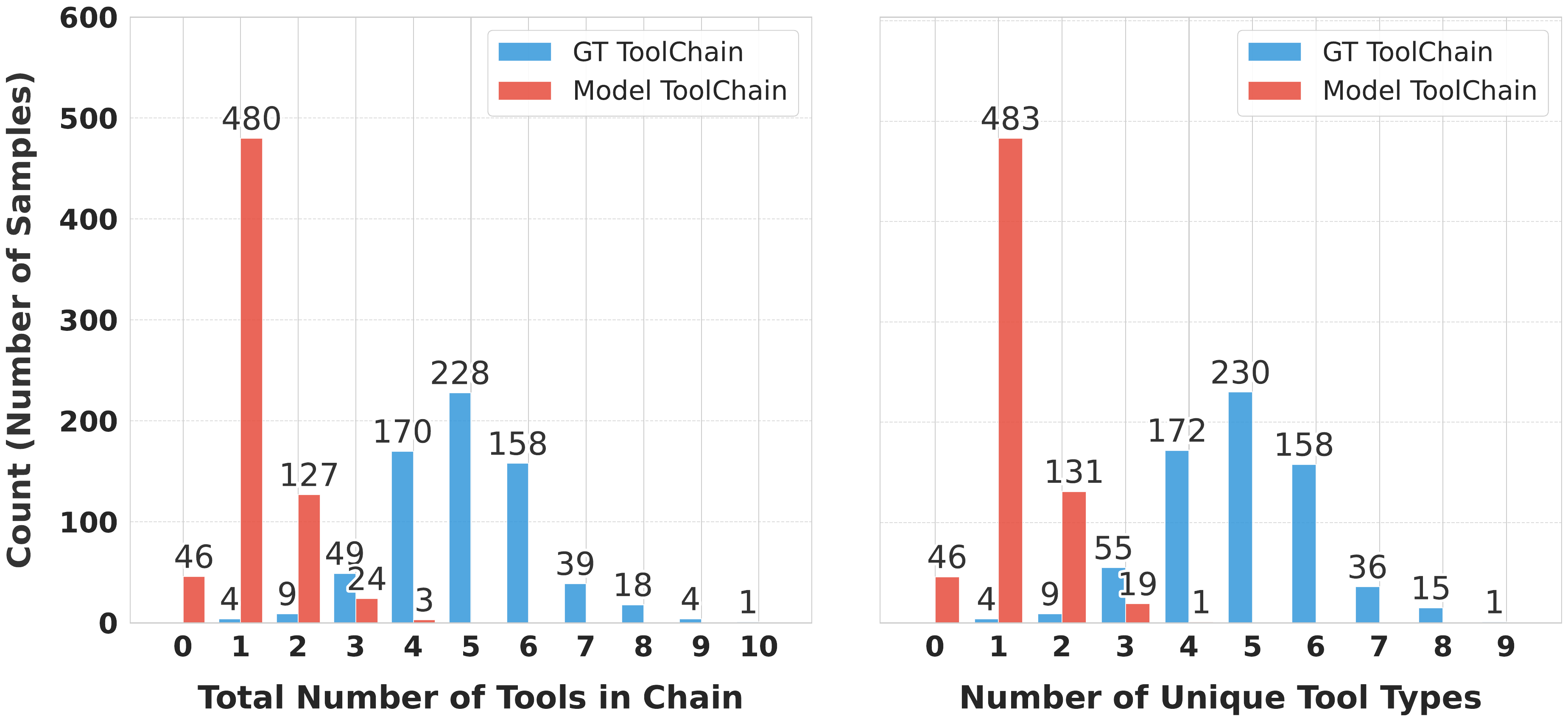}
        \caption{\textbf{Distributional Discrepancy.}}
        \label{fig:distribution_diff}
    \end{subfigure}
    
    \caption{\textbf{Comprehensive analysis for Gemini-3.0-Flash.} \textbf{(a)} Correlation between sample density and APR. \textbf{(b)} Distributional gap between Ground Truth  and Model. Both metrics are stratified by chain length (left) and tool diversity (right).}
    \label{fig:overall_analysis}
    \vspace{-0.5cm}
\end{figure}

\noindent \textbf{Tool composition enhances performance but remains a fundamental execution bottleneck.} Visualized in Fig.~\ref{fig:tool_analysis}, multi-round combinations positively correlate with overall performance, with trajectories exceeding two steps consistently surpassing average accuracy. However, distributional discrepancies (Fig.~\ref{fig:distribution_diff} and Fig.~\ref{fig:category_tool_distribution}) alongside the quantitative metrics in Tab.~\ref{tab:tool_analysis} expose a striking inefficiency. Paradoxically, even models with high APR fail to align with optimal, ground-truth tool-calling patterns. For example, despite its advanced reasoning capabilities, GPT-5.2 records a MAE of 9.96 in attempted calls and an Eff. of 16.78\%. Similarly, GPT-o3 and Gemini-3.0-Pro only achieve efficiencies of 17.20\% and 36.51\%, respectively. These pronounced deviations and excessive redundancies demonstrate that instead of executing precise programmatic orchestration, current models resort to suboptimal trial-and-error heuristics, exposing a critical deficit in coordinating complex, multi-step tool sequences.

\begin{figure}[t]
    \centering
    \begin{subfigure}{0.49\linewidth}
        \centering
        \includegraphics[width=\linewidth]{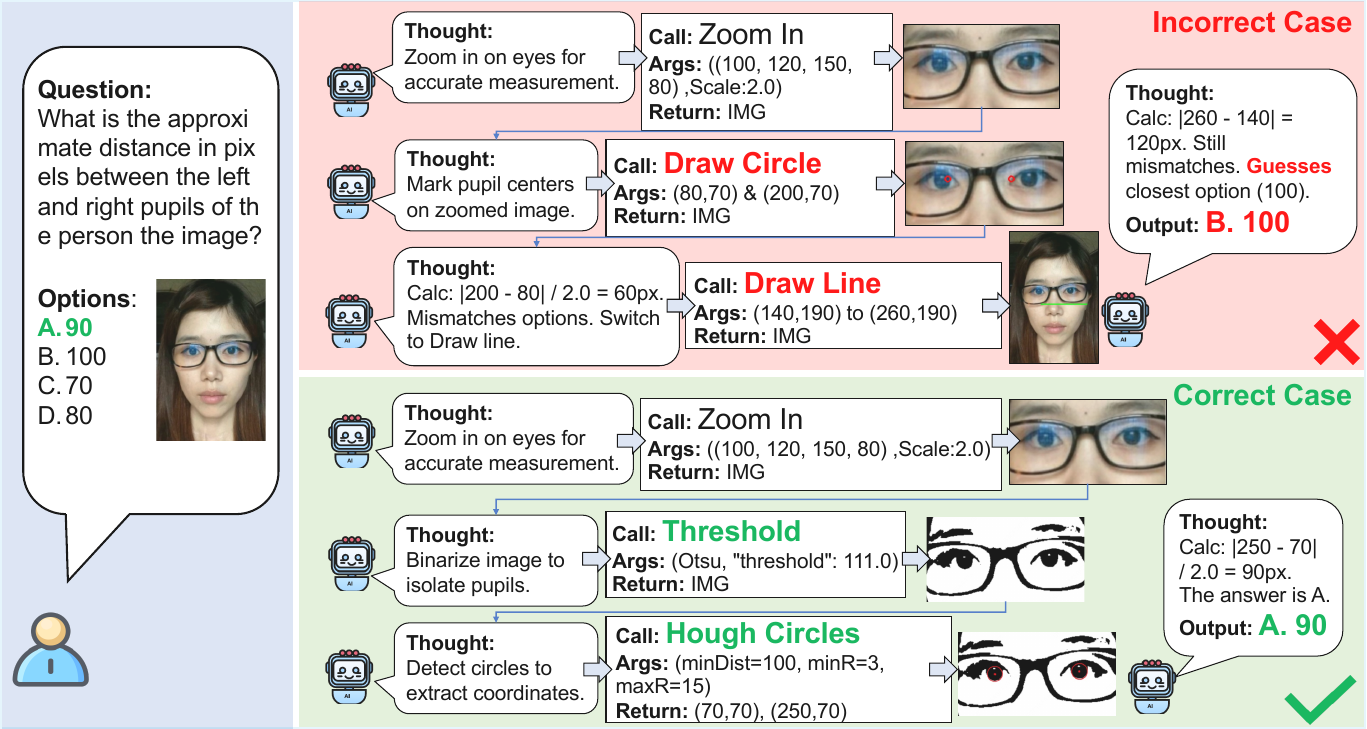}
        \caption{Tool selection and execution errors.}
        \label{fig:Bad_Case_1}
    \end{subfigure}
    \hfill 
    \begin{subfigure}{0.49\linewidth}
        \centering
        \includegraphics[width=\linewidth]{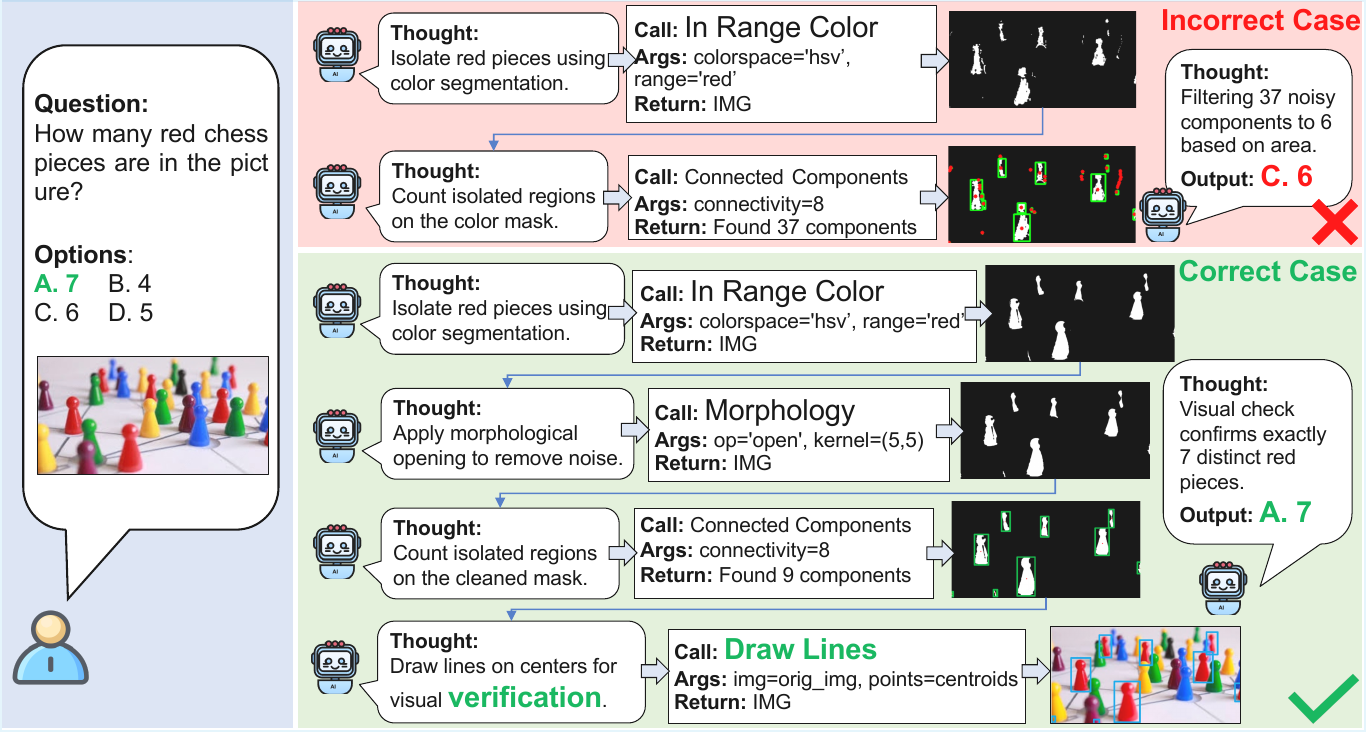}
        \caption{Over-reliance on unverified tool outputs.}
        \label{fig:Bad_Case_2}
    \end{subfigure}
    
    \caption{\textbf{Qualitative analysis of failure modes in tool utilization.} 
    (a) The agent mistakenly invokes \textit{Draw Circle} and \textit{Draw Line} based on unfounded intuition, leading to a speculative answer. 
    (b) The agent blindly adopts preliminary tool outputs without visual validation, bypassing essential verification steps.}
    \label{fig:Error_Analysis_Combined}
    
\end{figure}

\noindent \textbf{Case Study}~\label{sec:case_study}
A qualitative analysis of model outputs reveals two primary failure modes in tool utilization. The first mode involves a compounded failure of strategic misselection and operational errors during tool invocation. As Fig.~\ref{fig:Bad_Case_1} illustrates, the model not only selects the \textit{Draw Circle} and \textit{Draw Line} tools—which are fundamentally inappropriate for the given context—but also executes them incorrectly. Specifically, it attempts to apply these tools based solely on its flawed intrinsic perception, entirely bypassing the prerequisite step of extracting accurate spatial coordinates. This highlights a critical deficiency in understanding specific tool functionalities and their proper application contexts. The second mode is an over-reliance on intermediate tool results. Fig.~\ref{fig:Bad_Case_2} demonstrates this behavior, where the model conducts a superficial analysis of the returned output and directly adopts it to form the final answer. By failing to cross-verify these intermediate results with the original visual input, the model completely misses the correct target. This lack of critical verification prevents models from identifying and correcting potential errors generated by the tools.

\begin{figure}[t]
    \centering
    \includegraphics[width=0.8\linewidth]{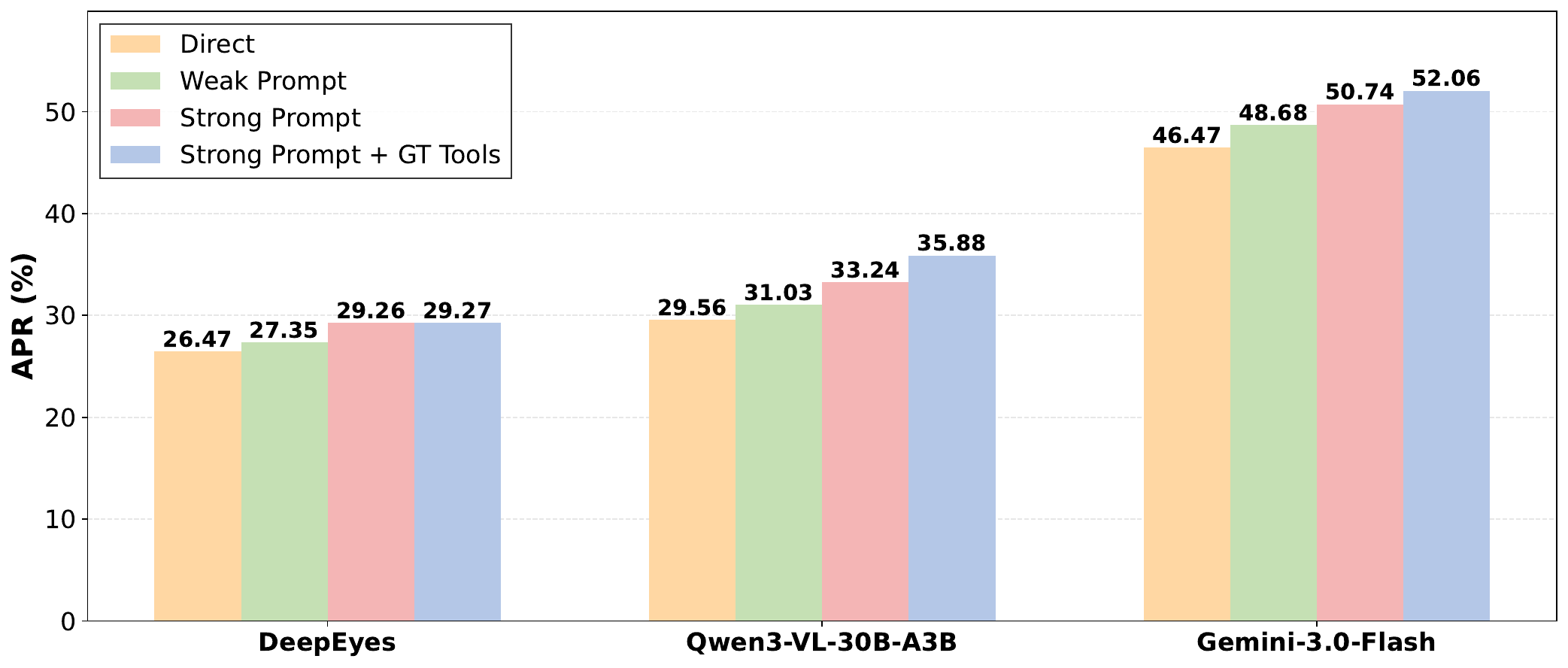}
    \caption{\textbf{Impact of system prompts and prompt information density on model APR.} We evaluate the performance of three distinct models, under four prompting configurations: Direct, Weak Prompt, Strong Prompt, and Strong Prompt + GT Tools.}
    \label{fig:prompt_ablation}
    \vspace{-0.5cm}
\end{figure}

\noindent \textbf{Prompt Ablation}~\label{sec:ablation_study}
We evaluate model performance across four distinct settings—direct answer, weak prompt, strong prompt (the default configuration for our main evaluations), and strong prompt augmented with ground-truth (GT) tools—with results detailed in Fig.~\ref{fig:prompt_ablation}. Our evaluations on Gemini-3.0-Flash~\cite{google2025gemini3flash}, Qwen3-VL-30B-A3B~\cite{qwen3-vl}, and DeepEyes~\cite{zheng2025deepeyes} reveal a consistent upward trajectory in performance correlating with prompt comprehensiveness. Notably, the performance improvements observed upon introducing GT tools effectively validate the utility and correctness of our annotated reference toolchains. Nevertheless, the overall gains remain surprisingly bounded even for leading models like Gemini and Qwen, while improvements for DeepEyes are entirely marginal. This performance ceiling suggests that providing the correct toolset is merely half the battle: restricted by their inherent reasoning capacities, current models fundamentally struggle to synthesize the correct multi-step execution logic. This inability to reliably formulate tool trajectories, even with oracle tool knowledge, underscores a profound bottleneck in complex compositional reasoning.


\section{Conclusion}

In this paper, we introduced \textbf{\benchname}, a meticulously crafted benchmark designed to evaluate the multi-step tool composition capabilities of MLLMs. 
By integrating 32 diverse tools across 9 real-world categories, our suite substantially surpasses prior evaluations. 
Extensive experiments across 19 MLLMs , including both open and proprietary, reveal pronounced deficiencies in current tool orchestration. 
Models frequently gravitate toward a narrow subset of tools, resulting in suboptimal execution efficacy. Moreover, models often struggle with precise tool invocation or demonstrate an over-reliance on tool outputs, neglecting their intrinsic perceptual capabilities. 
These findings underscore a critical disconnect between foundational visual perception and active agentic reasoning, illuminating a clear frontier for future MLLM architectures.


%
%
\bibliographystyle{splncs04}
\bibliography{main}

\newpage
\appendix

\onecolumn

\section{Implementation Details}
We conduct all evaluations in a zero-shot setting to ensure fair comparison and better generalization. Open-source models are deployed on NVIDIA H100 GPUs using the \texttt{vLLM} framework, while proprietary models are accessed via their official APIs. The detailed generation hyperparameters are set as follows:

\begin{itemize}
    \item \textbf{Open-source Instruct Models:} temperature $= 0.7$, top-p $= 0.8$, top-k $= 20$, repetition penalty $= 1.0$, presence penalty $= 1.5$, max tokens $= 16{,}384$, and seed $= 3407$.
    
    \item \textbf{Open-source Thinking Models:} temperature $= 0.6$, top-p $= 0.95$, top-k $= 20$, repetition penalty $= 1.0$, presence penalty $= 0$, max tokens $= 40{,}960$, and seed $= 1234$.
    
    \item \textbf{Proprietary Models:} We strictly enable the High Reasoning mode for all assessments and set the max new tokens to $65{,}536$.
\end{itemize}

In Qwen-Agent, we set the maximum number of rounds to 20.

\section{Detailed Tool Set And Data Collection}
\subsection{Tool Set}\label{app:tool_set}
All tools and their corresponding descriptions are detailed in Tab.~\ref{tab:tool_taxonomy}.
\begin{table}[tb]
    \centering
    \caption{Detailed taxonomy and definitions of the 35 OpenCV-based tools utilized in our benchmark. The tools are categorized into four logical groups: Geometry, Enhancement, Feature Extraction, and Drawing.}
    \label{tab:tool_taxonomy}
    
    \scriptsize 
    \renewcommand{\arraystretch}{1.2} 
    \setlength{\tabcolsep}{6pt} 
    
    \begin{tabularx}{\textwidth}{>{\bfseries}l X} 
        \toprule
        \rowcolor{white}
        Tool Name & \textbf{Description} \\
        \midrule
        
        \rowcolor{gray!15} \multicolumn{2}{c}{\textbf{\textsc{Geometry}}} \\
        Resize & Adjusts the image dimensions to a specified size. \\
        Rotate & Rotates the image by a specified angle. \\
        Translate & Shifts the image location along the x or y axis. \\
        Flip & Mirrors the image horizontally, vertically, or both. \\
        Crop & Extracts a specific rectangular region of interest (ROI). \\
        Zoom In & Scales up a specific region of the image for detail inspection. \\
        Pyramid & Performs pyramid upsampling or downsampling. \\
        
        \rowcolor{gray!15} \multicolumn{2}{c}{\textbf{\textsc{Enhancement}}} \\
        Convert Color & Converts color spaces (e.g., RGB to Gray, HSV, or LAB). \\
        In-Range Color & Filters pixels within a specific color range (segmentation mask). \\
        Blur & Smooths image noise using Gaussian, Median, or Bilateral filters. \\
        Denoise & Applies Non-local Means Denoising to smooth while preserving edges. \\
        Threshold & Applies binary, Otsu, or adaptive thresholding techniques. \\
        Morphology & Performs morphological operations (Erosion, Dilation, Open, Close). \\
        Histogram & Enhances contrast using Histogram Equalization or CLAHE. \\
        Adjust Brightness & Adjusts brightness using absolute scaling (ConvertScaleAbs). \\
        Inpaint & Restores damaged or missing areas within an image. \\
        
        \rowcolor{gray!15} \multicolumn{2}{c}{\textbf{\textsc{Feature Extraction}}} \\
        Canny Edge & Identifies edges using the Canny edge detector. \\
        Compute Gradients & Calculates gradients using Sobel or Laplacian operators. \\
        Watershed & Segments touching objects based on topographical distance. \\
        GrabCut & Extracts foreground objects interactively using graph cuts. \\
        Flood Fill & Fills a connected component starting from a seed point. \\
        Conn. Components & Computes connected components and their statistics (bounding box, area). \\
        Keypoint Features & Detects feature points using algorithms like Harris, SIFT, or ORB. \\
        Hough Lines & Detects straight lines using the Hough Line Transform. \\
        Hough Circles & Detects circular objects using the Hough Circle Transform. \\
        Template Match & Locates the area in the image that matches a template. \\
        DFT & Converts the image to the frequency domain (Discrete Fourier Transform). \\
        
        \rowcolor{gray!15} \multicolumn{2}{c}{\textbf{\textsc{Drawing}}} \\
        Draw Contours & Detects and retrieves contours from the binary image. \\
        Approx Poly & Approximates a contour shape to a polygon with fewer vertices. \\
        Draw Line & Annotates the image with straight lines. \\
        Draw Circle & Annotates the image with circles. \\
        Contour Area & Calculates the area of a specific contour. \\
        Arc Length & Computes the perimeter or curve length of a contour. \\
        \bottomrule
    \end{tabularx}
\end{table}

\subsection{Data Collection}\label{app:data_collection}
We describe the specific data collection process for the nine tasks in this section.

\textit{Attention Focusing:} Sourced from HRBench~\cite{wang2025divide} and web-crawled images. We evaluate the agent’s robustness against Geometric Perturbations by applying manual transformations (e.g., rotation, reflection). This requires models to re-orient their "focus" via tool-assisted spatial normalization.

\textit{Chart:} Derived from ChartQA~\cite{masry2022chartqa} and targeted web searches. We implement information sanitization by erasing original ground-truth labels, compelling the agent to utilize Auxiliary Construction for logic and Precision Sampling for element extraction.

\textit{Color:} Based on ColorBench~\cite{liang2025colorbench} and internet data. After subjecting images to radiometric noise (blur, illumination variance), we force the model to bypass raw RGB perception in favor of chromatic space manipulations (HSV/LAB) to quantify color proportions.

\textit{Counting:} Adapted from Kaggle and web sources. We specifically target visual occlusion and density bottlenecks (e.g., overlapping objects) where direct counting is intractable, necessitating a "segment-and-count" pipeline using morphological utilities.

\textit{Math:} A specialized set of web-collected and manually synthesized samples focusing on STEM-oriented Geometric Reasoning. The core challenge lies in constructive logic, where models must identify and draw auxiliary lines to solve complex polygonal problems.

\textit{Measurement:} Sourced from high-fidelity web images with manual calibration. It bridges the gap between vision and Metrology, featuring scenarios like Industrial Component Calibration that require sub-pixel precision through tool-assisted physical dimension estimation.

\textit{Perceptual Restoration:} Leveraging Kaggle open-source data, this task evaluates the agent’s capacity to act as a "pre-processor," effectively neutralizing atmospheric haze and photon noise to recover latent semantic information from degraded scenes.

\textit{Robust OCR:} An integration of OCRBench~\cite{liu2024ocrbench} and Kaggle competition data. Unlike standard OCR, these samples are embedded with compound degradations. Models must exhibit strategic planning by chaining binarization and sharpening tools before attempting text recognition. 

\textit{Spatial Reasoning}: Derived from Kaggle and web sources. This task focuses on the quantification of relative topologies, requiring models to transform qualitative visual cues into precise spatial coordinates to resolve complex positional queries.

\section{More On Analysis Results}
\subsection{Tool-Use Analysis Results}
\begin{figure}[t]
    \centering
    \includegraphics[width=0.95\linewidth]{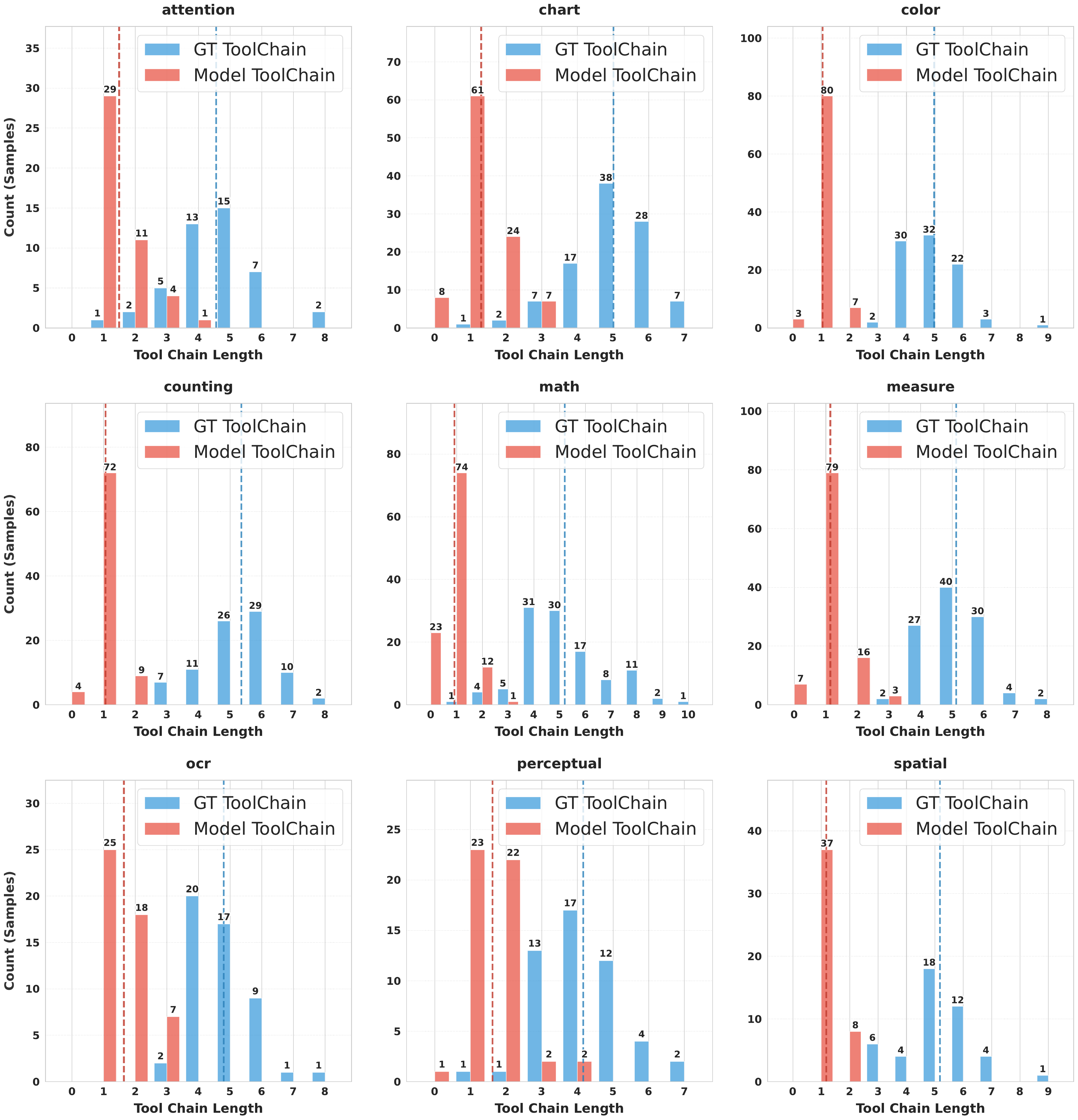}
    \caption{Category-level distribution difference compared with ground truth for Gemini-3.0-Flash.}
    \label{fig:category_tool_distribution}
\end{figure}

Figure~\ref{fig:category_tool_distribution} visualizes the distribution difference of toolchain lengths between Gemini-3.0-Flash and the ground truth at a granular task level. The results demonstrate a critical mismatch between predicted toolchain depths and ground-truth requirements across all task categories. The model consistently generates significantly shorter toolchains than those required by expert trajectories. Most predicted sequences peak at only 1 or 2 steps while the ground truth frequently requires 4 to 6 steps. In categories such as color and measure, the model produces a single-step chain for the vast majority of samples despite the ground truth peaking at 5 steps. This "shortcutting" behavior suggests that the model often terminates its reasoning process prematurely without completing the necessary intermediate steps. Such a depth deficiency identifies a major bottleneck in the planning and orchestration capabilities of current large multi-modal models.

\subsection{Evaluation Metrics}\label{app: metrics}

We define the Tool Call Rate (TCR) as Eq~\ref{Eq: TCR}:
\begin{equation}\label{Eq: TCR}
 \text{TCR} = \frac{|\{i \mid L_{T,i} > 0\}|}{N},
\end{equation}
where $N$ denotes the total number of samples and $L_{T,i}$ represents the toolchain length for the $i$-th sample.

\clearpage
\section{Detailed Prompts}\label{app:prompts}
\begin{tcolorbox}[
    enhanced,               
    breakable,
    colback=bluebg,         
    colframe=blueframe,     
    coltitle=white,         
    title={System Prompt},  
    fonttitle=\bfseries\sffamily, 
    fontupper=\footnotesize,
    arc=2.5mm,              
    boxrule=1.2pt,          
    drop fuzzy shadow,      
    left=10pt, right=10pt, top=10pt, bottom=10pt 
]

\noindent\textbf{\textsf{System Prompt (Strong)}}\\
\vspace{1mm}

\noindent Your role is that of a helpful assistant specialized in solving real-world visual problems using image processing tools. Answer questions about images by combining your visual understanding with the precision of available tools.

\noindent Please follow this structured thinking process and show your work.

\noindent Start an iterative loop for each question:

\noindent - **First, look closely:** Begin with a detailed description of the image in the context of the user's real-life query. List what is immediately visible, and explicitly identify what specific visual details need to be clarified, adjusted, or measured using tools.

\noindent - **Next, apply tools:** Select and invoke just one appropriate function to process the image.

\noindent - **Then, review the findings:** Carefully analyze the tool's output (e.g., the processed image, detected features, or status) and decide on your next action (e.g., did the rotation fix the view? do you need further adjustments?).

\noindent Continue this loop until you have sufficient information.

\noindent To finish, bring everything together in a clear, synthesized answer that fully responds to the user's question.

\vspace{2mm}
\tcbline 
\vspace{2mm}

\noindent\textbf{\textsf{System Prompt (Weak)}}\\
\vspace{1mm}

\noindent Your role is that of a helpful assistant specialized in solving real-world visual problems using image processing tools. 

\end{tcolorbox}

\begin{tcolorbox}[
    enhanced,               
    breakable,
    colback=bluebg,         
    colframe=blueframe,     
    coltitle=white,         
    title={User Prompt},  
    fonttitle=\bfseries\sffamily, 
    fontupper=\footnotesize,
    arc=2.5mm,              
    boxrule=1.2pt,          
    drop fuzzy shadow,      
    left=10pt, right=10pt, top=10pt, bottom=10pt 
]

\noindent\textbf{\textsf{User Prompt without GT Toolchains}}\\
\vspace{1mm}

\noindent   <image>
  
\noindent  \{question\}
  
\noindent  \#\#\# User Image Path: \{image\_path\}
  
\noindent  \#\#\#  User Image Size: \{image\_size\}
  
\noindent  \#\#\#  Output Format (strict adherence required):
  
\noindent  <think>Your detailed reasoning process, should go here.</think>
  
\noindent  <answer>Your final answer to the user's question goes here.</answer>

\vspace{2mm}
\tcbline 
\vspace{2mm}

\noindent\textbf{\textsf{User Prompt with GT Toolchains}}\\
\vspace{1mm}

\noindent   <image>
  
\noindent  \{question\}
  
\noindent  \#\#\# User Image Path: \{image\_path\}
  
\noindent  \#\#\#  User Image Size: \{image\_size\}

\noindent \#\#\# \textbf{
 Here is the **Reference Trajectory**. You can refer to this trajectory to use tools to solve problems :  
\{reference\_trajectory\}
}

\noindent  \#\#\#  Output Format (strict adherence required):
  
\noindent  <think>Your detailed reasoning process, should go here.</think>
  
\noindent  <answer>Your final answer to the user's question goes here.</answer>

\end{tcolorbox}

\begin{tcolorbox}[
    enhanced,               
    breakable,
    colback=bluebg,         
    colframe=blueframe,     
    coltitle=white,         
    title={Code Interpreter Description},  
    fontupper=\footnotesize,             
    fonttitle=\bfseries\sffamily, 
    arc=2.5mm,              
    boxrule=1.2pt,          
    drop fuzzy shadow,      
    left=10pt, right=10pt, top=10pt, bottom=10pt 
]

\noindent Executes Python code to handle images based strictly on the allowed capabilities listed below.

\vspace{4mm}
\noindent CRITICAL INSTRUCTIONS FOR MODEL:
\begin{enumerate}
    \item RESTRICTED SCOPE: You are strictly permitted to implement only the specific image handling operations defined in the Allowed Capabilities Reference section below. Do not generate code for operations outside this list.
    \item NO PRE-DEFINED FUNCTIONS: The list below defines behaviors, NOT callable functions.
    \begin{itemize}
        \item INCORRECT: Calling colorspace\_gray(image) or resize(image, param). These functions DO NOT exist.
        \item CORRECT: You must write the raw Python code using the cv2 library to implement the logic described. For example, to achieve colorspace\_gray, you must write cv2.cvtColor(image, cv2.COLOR\_BGR2GRAY).
    \end{itemize}
    \item PARAMETER ADHERENCE: When writing the code, you must strictly follow the parameter logic described in the reference list (e.g., input ranges, default values, and calculation logic).
\end{enumerate}

\vspace{4mm}
\tcbline
\vspace{4mm}

\noindent Allowed Capabilities Reference (Implement these using raw cv2 code)

\vspace{2mm}

\noindent 1. COLOR SPACE CONVERSION TOOLS
\vspace{2mm}

\noindent \textbf{colorspace\_gray}
\newline Description: Converts the image to grayscale color space. Useful for enhancing contrast or isolating intensity information.
\newline Parameters:
\begin{itemize}
    \item image (string, required): The input image identifier
    \item param (object, optional): Empty parameter object
\end{itemize}

\noindent \textbf{colorspace\_hsv}
\newline Description: Converts the image to HSV (Hue, Saturation, Value) color space. Useful for color-based segmentation or isolating specific color channels.
\newline Parameters:
\begin{itemize}
    \item image (string, required): The input image identifier
    \item param (object, optional): Empty parameter object
\end{itemize}

\noindent \textbf{colorspace\_lab}
\newline Description: Converts the image to LAB color space. Useful for perceptual color operations or isolating specific color channels.
\newline Parameters:
\begin{itemize}
    \item image (string, required): The input image identifier
    \item param (object, optional): Empty parameter object
\end{itemize}

\vspace{4mm}
\noindent 2. GEOMETRIC TRANSFORMATION TOOLS
\vspace{2mm}

\noindent \textbf{resize}
\newline Description: Resizes the image to specified dimensions or by a preset scale (half or double).
\newline Parameters:
\begin{itemize}
    \item image (string, required): The input image identifier
    \item param (object, required): width (integer), height (integer), preset (string, optional)
\end{itemize}

\noindent \textbf{rotate}
\newline Description: Rotates the image by specified angle in degrees (clockwise).
\newline Parameters:
\begin{itemize}
    \item image (string, required): The input image identifier
    \item param (object, required): angle (number, required)
\end{itemize}

\noindent \textbf{translate}
\newline Description: Shifts the image by a specified number of pixels in the specified direction.
\newline Parameters:
\begin{itemize}
    \item image (string, required): The input image identifier
    \item param (object, required): direction (string, optional), distance (integer, optional)
\end{itemize}

\noindent \textbf{flip}
\newline Description: Flips the image horizontally or vertically.
\newline Parameters:
\begin{itemize}
    \item image (string, required): The input image identifier
    \item param (object, required): direction (string, optional)
\end{itemize}

\noindent \textbf{crop}
\newline Description: Crops a rectangular region from the image.
\newline Parameters:
\begin{itemize}
    \item image (string, required): The input image identifier
    \item param (object, optional): x (integer), y (integer), width (integer), height (integer)
\end{itemize}

\noindent \textbf{zoom\_in}
\newline Description: Zooms into a specific region of the image by cropping and optionally resizing.
\newline Parameters:
\begin{itemize}
    \item image (string, required): The input image identifier
    \item param (object, optional): x, y, width, height, scale, target\_width, target\_height
\end{itemize}

\vspace{4mm}
\noindent 3. FILTERING AND SMOOTHING TOOLS
\vspace{2mm}

\noindent \textbf{blur}
\newline Description: Applies blurring to reduce noise or smooth the image. Supports average, gaussian, median, and bilateral.
\newline Parameters:
\begin{itemize}
    \item image (string, required): The input image identifier
    \item param (object, required): method (string), ksize (integer), sigma\_x, sigma\_y, d, sigma\_color, sigma\_space
\end{itemize}

\noindent \textbf{denoise}
\newline Description: Applies fast non-local means denoising.
\newline Parameters:
\begin{itemize}
    \item image (string, required): The input image identifier
    \item param (object, optional): mode, h, h\_color, template\_window, search\_window, channels
\end{itemize}

\vspace{4mm}
\noindent 4. THRESHOLDING AND BINARIZATION TOOLS
\vspace{2mm}

\noindent \textbf{threshold}
\newline Description: Applies thresholding to create a binary image.
\newline Parameters:
\begin{itemize}
    \item image (string, required): The input image identifier
    \item param (object, required): mode, invert, color\_mode, threshold\_value, adaptive\_block\_size, adaptive\_constant, channels
\end{itemize}

\noindent \textbf{inrange\_color}
\newline Description: Creates a mask for pixels within a specified color range.
\newline Parameters:
\begin{itemize}
    \item image (string, required): The input image identifier
    \item param (object, optional): colorspace, lower, upper, output\_format
\end{itemize}

\vspace{4mm}
\noindent 5. MORPHOLOGICAL OPERATIONS
\vspace{2mm}

\noindent \textbf{morphology}
\newline Description: Applies morphological operations (erode, dilate, open, close).
\newline Parameters:
\begin{itemize}
    \item image (string, required): The input image identifier
    \item param (object, optional): op, kernel\_size, iterations, kernel\_shape
\end{itemize}

\vspace{4mm}
\noindent 6. EDGE DETECTION TOOLS
\vspace{2mm}

\noindent \textbf{gradients}
\newline Description: Computes gradient images using Sobel or Laplacian operator.
\newline Parameters:
\begin{itemize}
    \item image (string, required): The input image identifier
    \item param (object, required): mode (sobel\_x, sobel\_y, laplacian)
\end{itemize}

\noindent \textbf{canny}
\newline Description: Detects edges using Canny edge detector.
\newline Parameters:
\begin{itemize}
    \item image (string, required): The input image identifier
    \item param (object, optional): preset, threshold\_low, threshold\_high, color\_mode, channels
\end{itemize}

\noindent \textbf{convertscaleabs}
\newline Description: Converts image to absolute value and scales it to 0-255 range.
\newline Parameters:
\begin{itemize}
    \item image (string, required): The input image identifier
    \item param (object, optional): alpha, beta
\end{itemize}

\vspace{4mm}
\noindent 7. CONTOUR DETECTION AND ANALYSIS TOOLS
\vspace{2mm}

\noindent \textbf{contours}
\newline Description: Finds contours using Canny edges and returns bounding boxes/areas.
\newline Parameters:
\begin{itemize}
    \item image (string, required): The input image identifier
    \item param (object, optional): mode, canny\_low, canny\_high, rank, max\_contours
\end{itemize}

\noindent \textbf{contour\_area}
\newline Description: Calculates contour areas using cv2.contourArea.
\newline Parameters:
\begin{itemize}
    \item image (string, required): The input image identifier
    \item param (object, optional): mode, canny\_low, canny\_high, rank, max\_contours, color, thickness
\end{itemize}

\noindent \textbf{arc\_length}
\newline Description: Computes contour perimeters using cv2.arcLength.
\newline Parameters:
\begin{itemize}
    \item image (string, required): The input image identifier
    \item param (object, optional): mode, canny\_low, canny\_high, rank, max\_contours, color, thickness
\end{itemize}

\noindent \textbf{approx\_poly}
\newline Description: Approximates contours with fewer points using cv2.approxPolyDP.
\newline Parameters:
\begin{itemize}
    \item image (string, required): The input image identifier
    \item param (object, optional): epsilon\_ratio, mode, canny\_low, canny\_high, rank, max\_contours, color, thickness
\end{itemize}

\noindent \textbf{connected\_components\_with\_stats}
\newline Description: Finds and analyzes connected components in a binary image.
\newline Parameters:
\begin{itemize}
    \item image (string, required): The input image identifier
    \item param (object, optional): threshold\_mode, threshold\_value, connectivity, max\_components
\end{itemize}

\vspace{4mm}
\noindent 8. SHAPE DETECTION TOOLS
\vspace{2mm}

\noindent \textbf{hough\_lines}
\newline Description: Detects line segments in the image using Hough transform.
\newline Parameters:
\begin{itemize}
    \item image (string, required): The input image identifier
    \item param (object, optional): threshold, minLineLength, maxLineGap, canny\_low, canny\_high, max\_lines
\end{itemize}

\noindent \textbf{hough\_circles}
\newline Description: Detects circles in the image using Hough transform.
\newline Parameters:
\begin{itemize}
    \item image (string, required): The input image identifier
    \item param (object, optional): dp, minDist, param1, param2, minRadius, maxRadius, max\_circles
\end{itemize}

\vspace{4mm}
\noindent 9. HISTOGRAM AND CONTRAST ENHANCEMENT
\vspace{2mm}

\noindent \textbf{histogram}
\newline Description: Applies histogram equalization or CLAHE to enhance image contrast.
\newline Parameters:
\begin{itemize}
    \item image (string, required): The input image identifier
    \item param (object, optional): mode, color\_mode, channels, clip\_limit, tile\_grid\_size
\end{itemize}

\vspace{4mm}
\noindent 10. IMAGE SEGMENTATION TOOLS
\vspace{2mm}

\noindent \textbf{watershed}
\newline Description: Applies watershed segmentation to separate overlapping objects.
\newline Parameters:
\begin{itemize}
    \item image (string, required): The input image identifier
    \item param (object, optional): max\_regions
\end{itemize}

\noindent \textbf{grabcut}
\newline Description: Performs foreground/background segmentation using GrabCut algorithm.
\newline Parameters:
\begin{itemize}
    \item image (string, required): The input image identifier
    \item param (object, required): preset (center, tight, loose)
\end{itemize}

\noindent \textbf{floodfill}
\newline Description: Performs flood fill operation starting from a seed point.
\newline Parameters:
\begin{itemize}
    \item image (string, required): The input image identifier
    \item param (object, optional): x, y, loDiff, upDiff, newVal, flags
\end{itemize}

\vspace{4mm}
\noindent 11. ADVANCED OPERATIONS
\vspace{2mm}

\noindent \textbf{pyramid}
\newline Description: Applies image pyramid operations (upsample or downsample).
\newline Parameters:
\begin{itemize}
    \item image (string, required): The input image identifier
    \item param (object, required): mode (pyr\_up, pyr\_down)
\end{itemize}

\noindent \textbf{dft}
\newline Description: Computes and visualizes the Discrete Fourier Transform magnitude spectrum.
\newline Parameters:
\begin{itemize}
    \item image (string, required): The input image identifier
    \item param (object, optional): Empty parameter object
\end{itemize}

\noindent \textbf{template\_match}
\newline Description: Matches a template image within the source image.
\newline Parameters:
\begin{itemize}
    \item image (string, required): The input image identifier
    \item param (object, optional): template\_path, method
\end{itemize}

\noindent \textbf{features}
\newline Description: Detects and draws keypoints using various feature detection methods.
\newline Parameters:
\begin{itemize}
    \item image (string, required): The input image identifier
    \item param (object, optional): method (harris, sift, orb, etc.), max\_points
\end{itemize}

\noindent \textbf{inpaint}
\newline Description: Inpaints (fills) regions in the image using automatically generated mask.
\newline Parameters:
\begin{itemize}
    \item image (string, required): The input image identifier
    \item param (object, optional): preset, canny\_low, canny\_high, threshold\_value, radius, method
\end{itemize}

\vspace{4mm}
\noindent 12. DRAWING TOOLS
\vspace{2mm}

\noindent \textbf{draw\_line}
\newline Description: Draws a line segment between two points.
\newline Parameters:
\begin{itemize}
    \item image (string, required): The input image identifier
    \item param (object, required): x1, y1, x2, y2, color, thickness
\end{itemize}

\noindent \textbf{draw\_circle}
\newline Description: Draws a circle on the image at specified center coordinates.
\newline Parameters:
\begin{itemize}
    \item image (string, required): The input image identifier
    \item param (object, optional): x, y, radius, color, thickness
\end{itemize}

\end{tcolorbox}

\clearpage
\section{Detailed Task Example}\label{app:detaled_task_example}
\begin{figure}[h]
    \centering
    \includegraphics[width=0.95\linewidth]{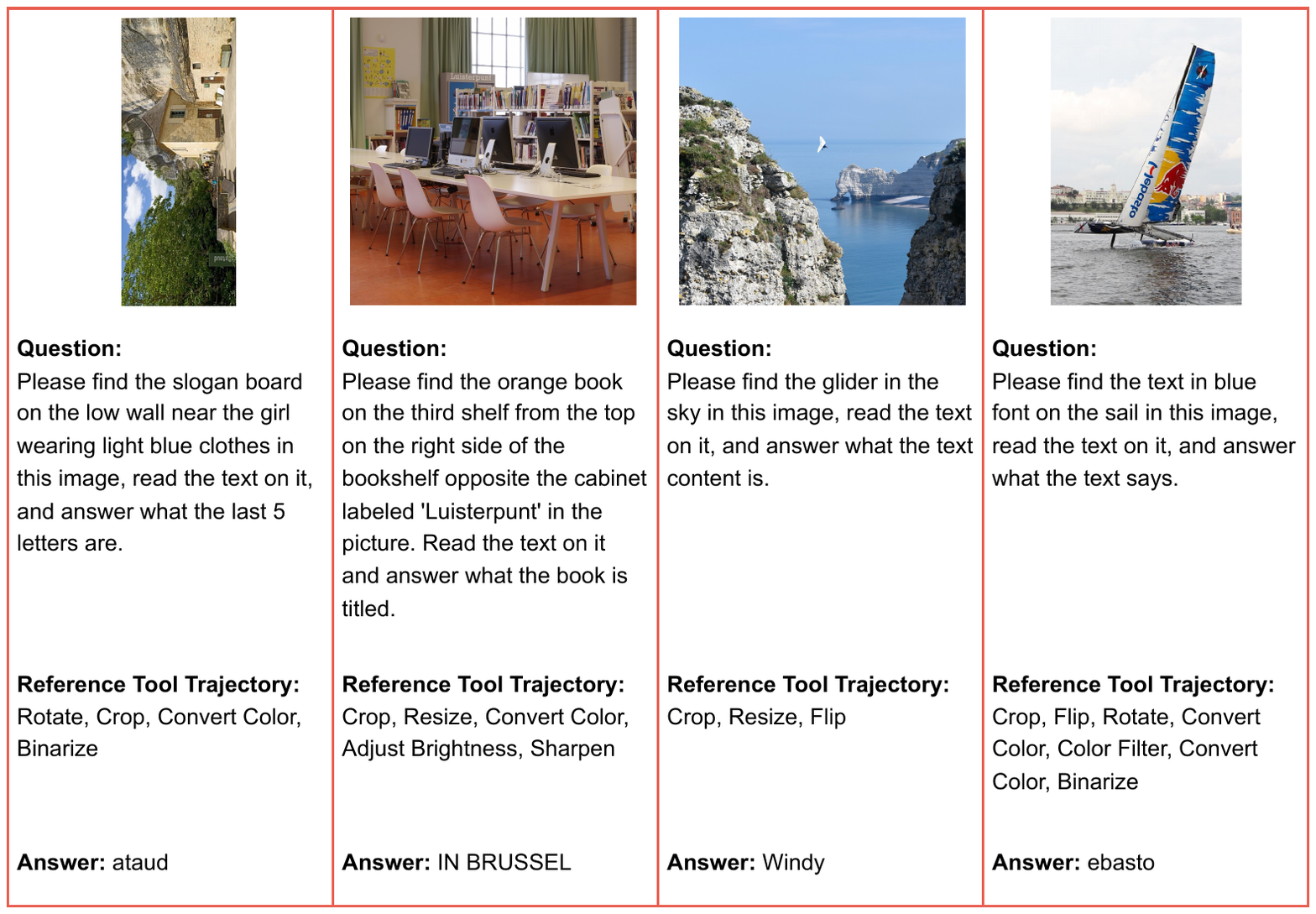}
    \caption{Supplementary examples of Attention Focusing tasks.}
    \label{fig:appendix_attention}
\end{figure}

\begin{figure}[t]
    \centering
    \includegraphics[width=0.95\linewidth]{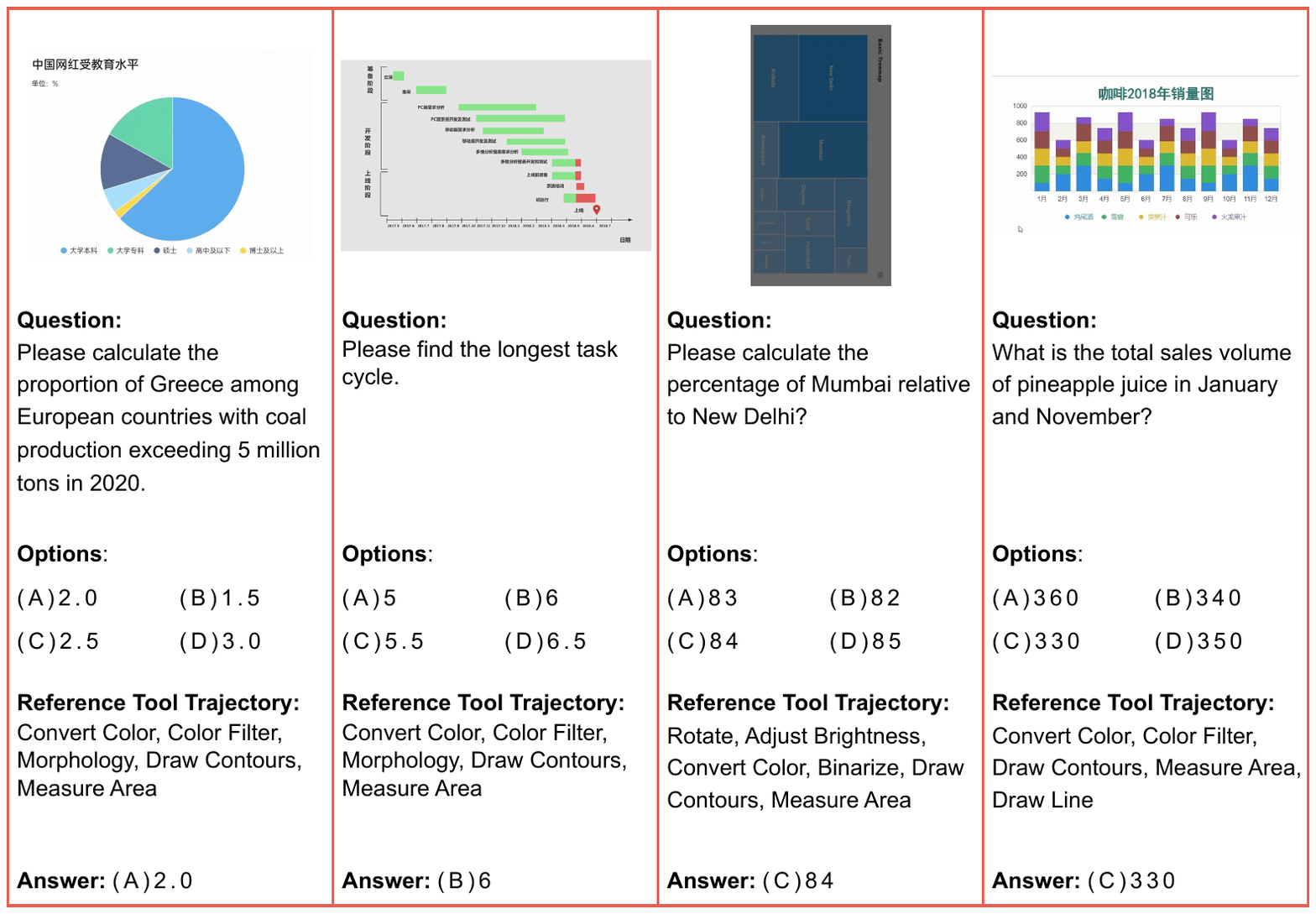}
    \caption{Supplementary examples of Chart tasks.}
    \label{fig:appendix_chart}
\end{figure}

\begin{figure}[t]
    \centering
    \includegraphics[width=0.95\linewidth]{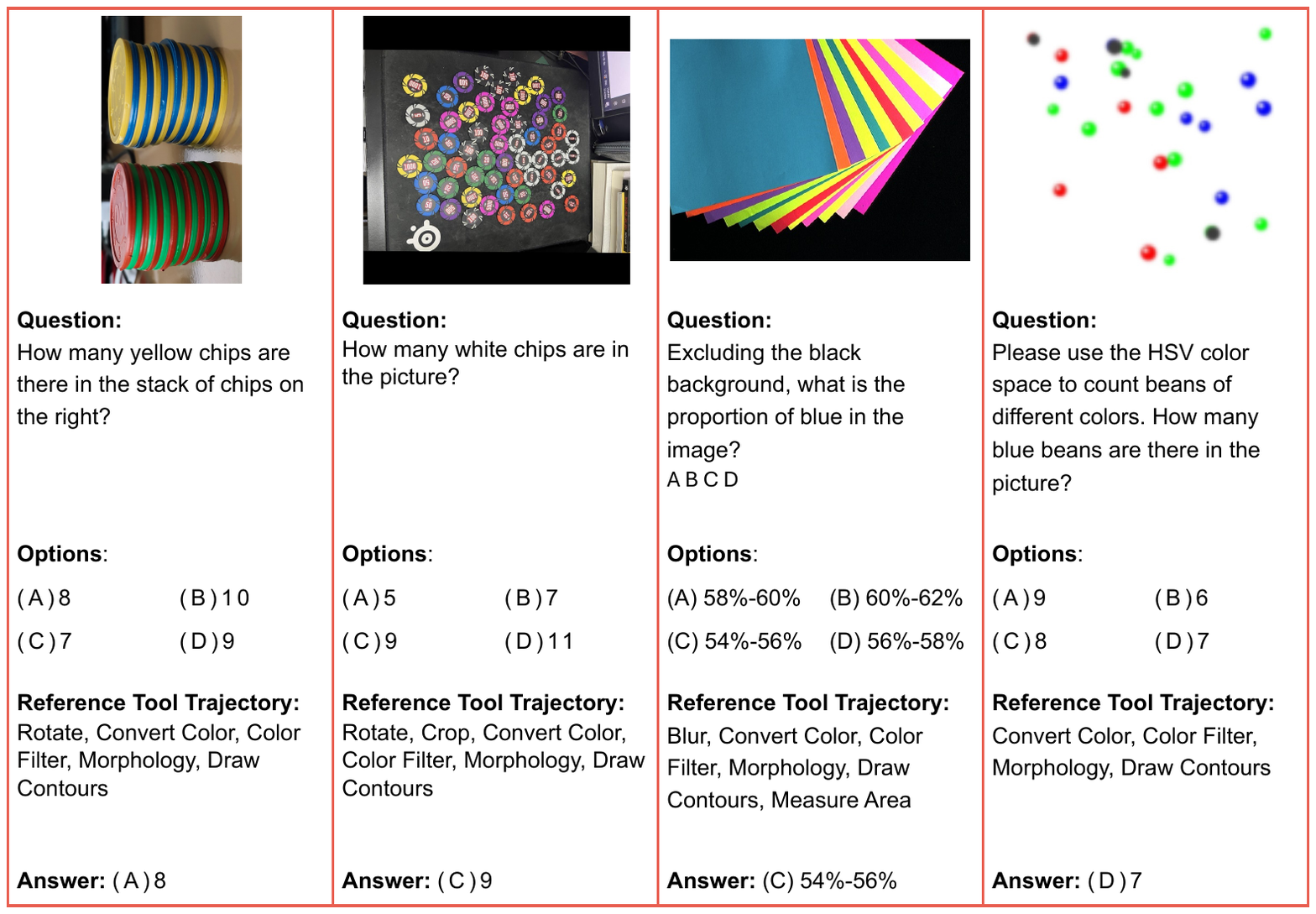}
    \caption{Supplementary examples of Color tasks.}
    \label{fig:appendix_color}
\end{figure}

\begin{figure}[t]
    \centering
    \includegraphics[width=0.95\linewidth]{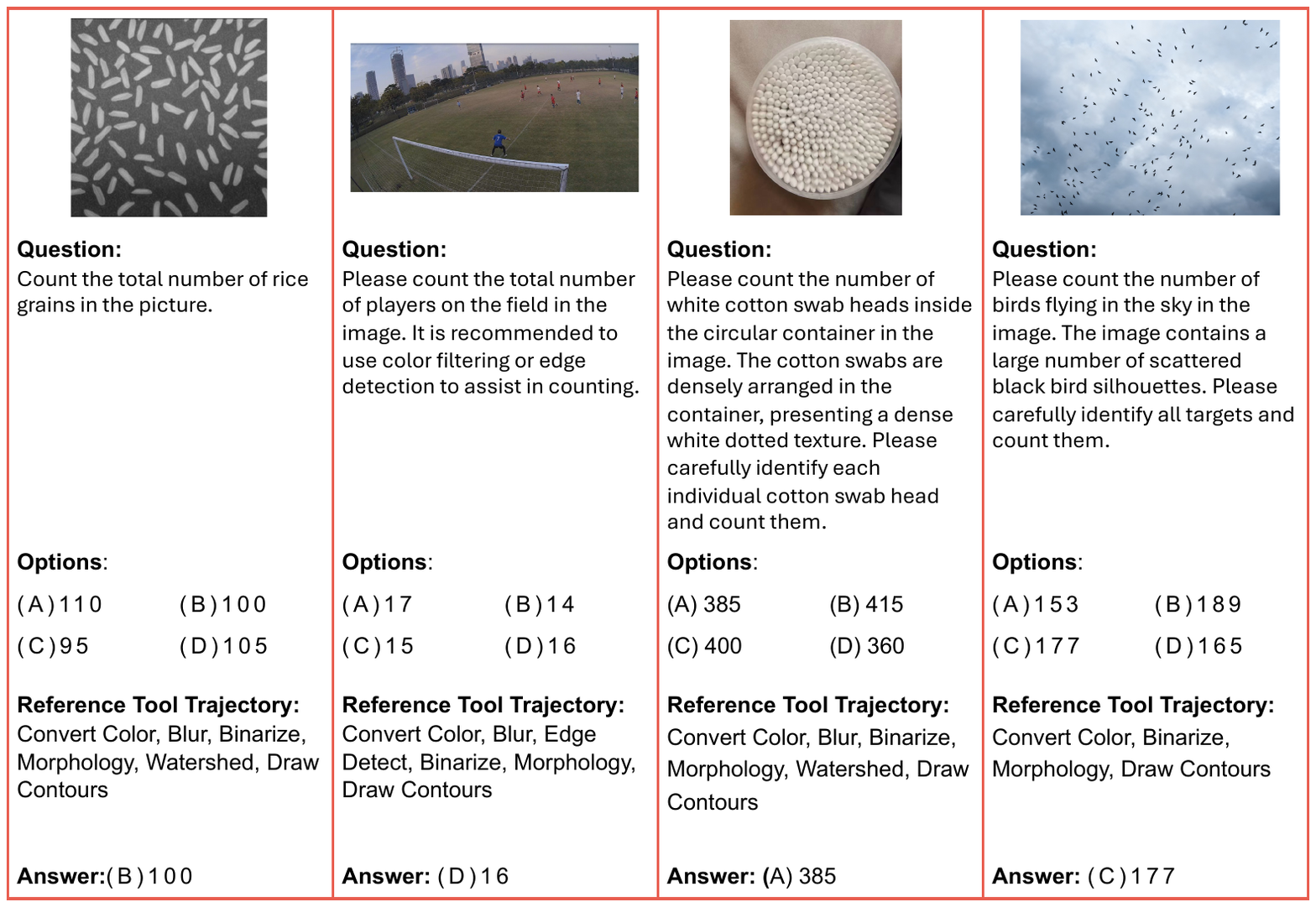}
    \caption{Supplementary examples of Counting tasks.}
    \label{fig:appendix_counting}
\end{figure}

\begin{figure}[t]
    \centering
    \includegraphics[width=0.95\linewidth]{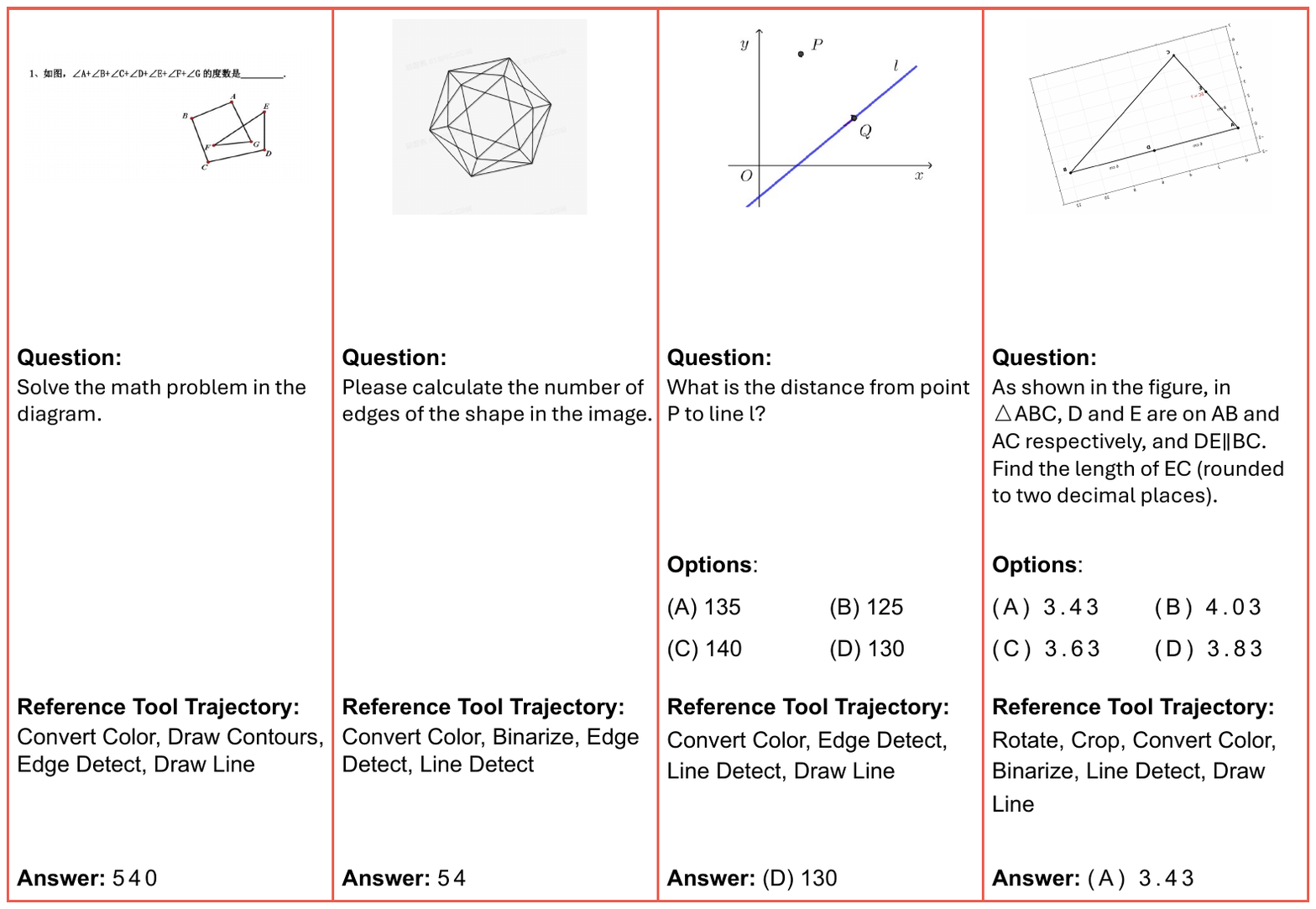}
    \caption{Supplementary examples of Math tasks.}
    \label{fig:appendix_math}
\end{figure}

\begin{figure}[t]
    \centering
    \includegraphics[width=0.95\linewidth]{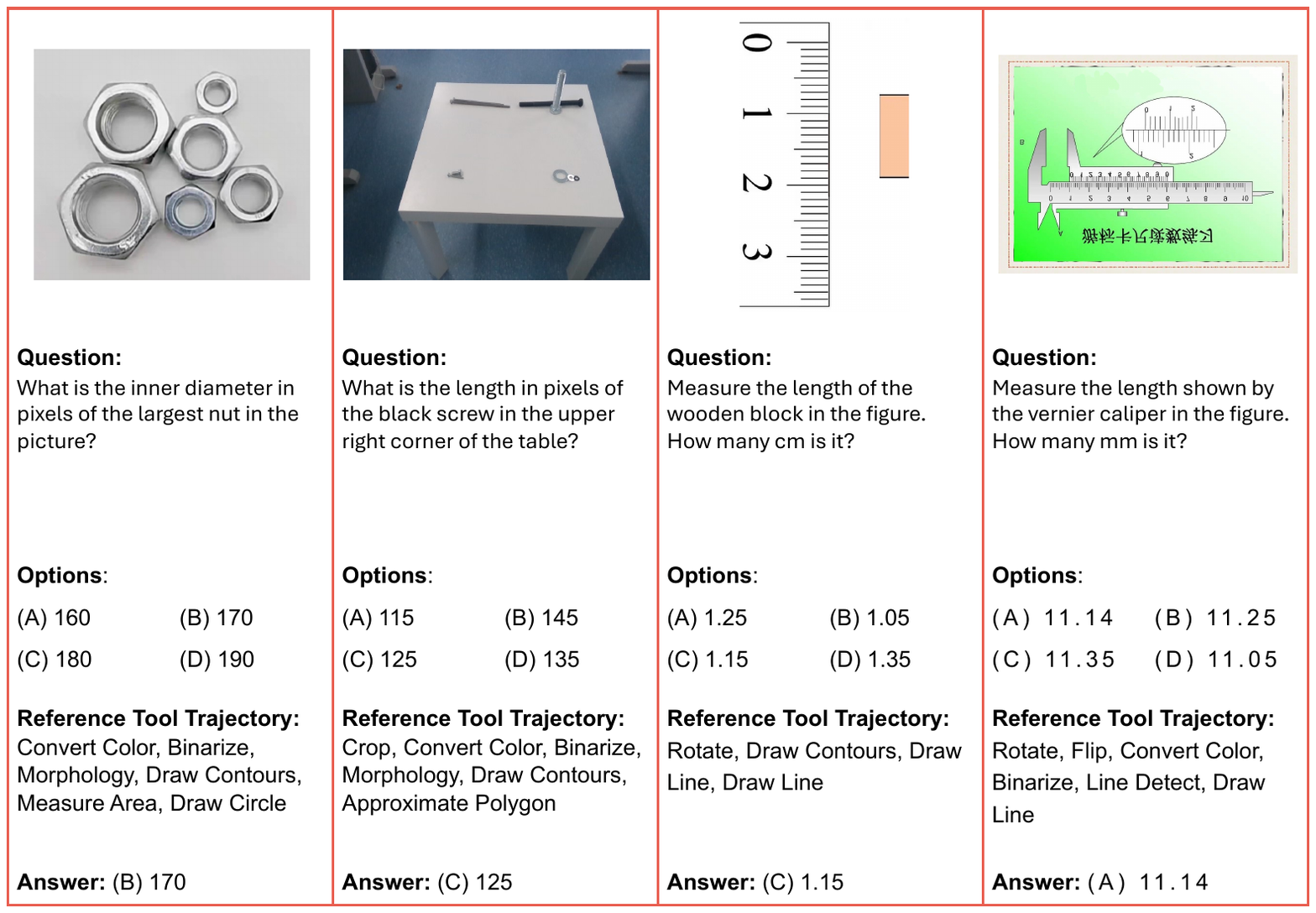}
    \caption{Supplementary examples of Measure tasks.}
    \label{fig:appendix_measure}
\end{figure}

\begin{figure}[t]
    \centering
    \includegraphics[width=0.95\linewidth]{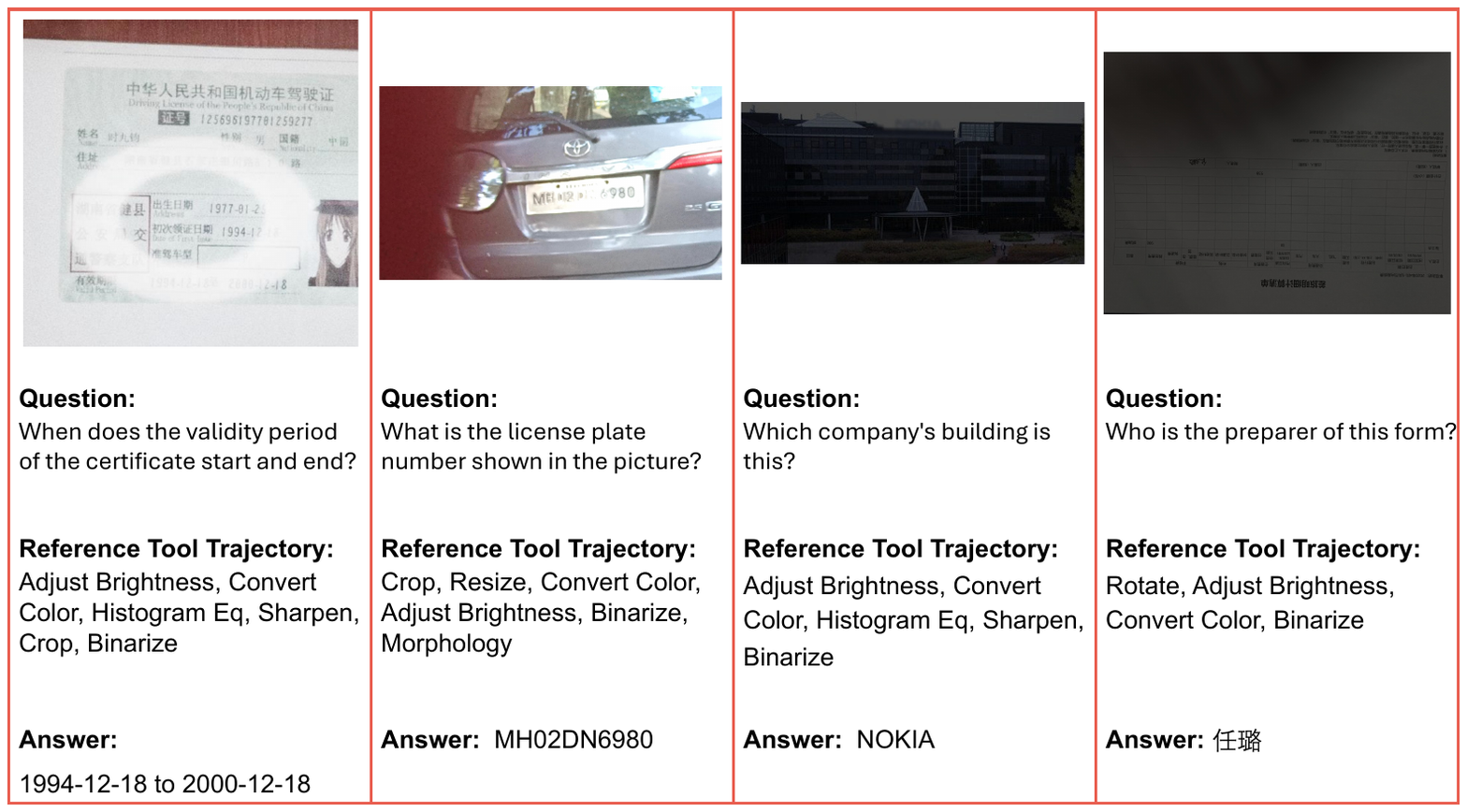}
    \caption{Supplementary examples of OCR tasks.}
    \label{fig:appendix_ocr}
\end{figure}

\begin{figure}[t]
    \centering
    \includegraphics[width=0.95\linewidth]{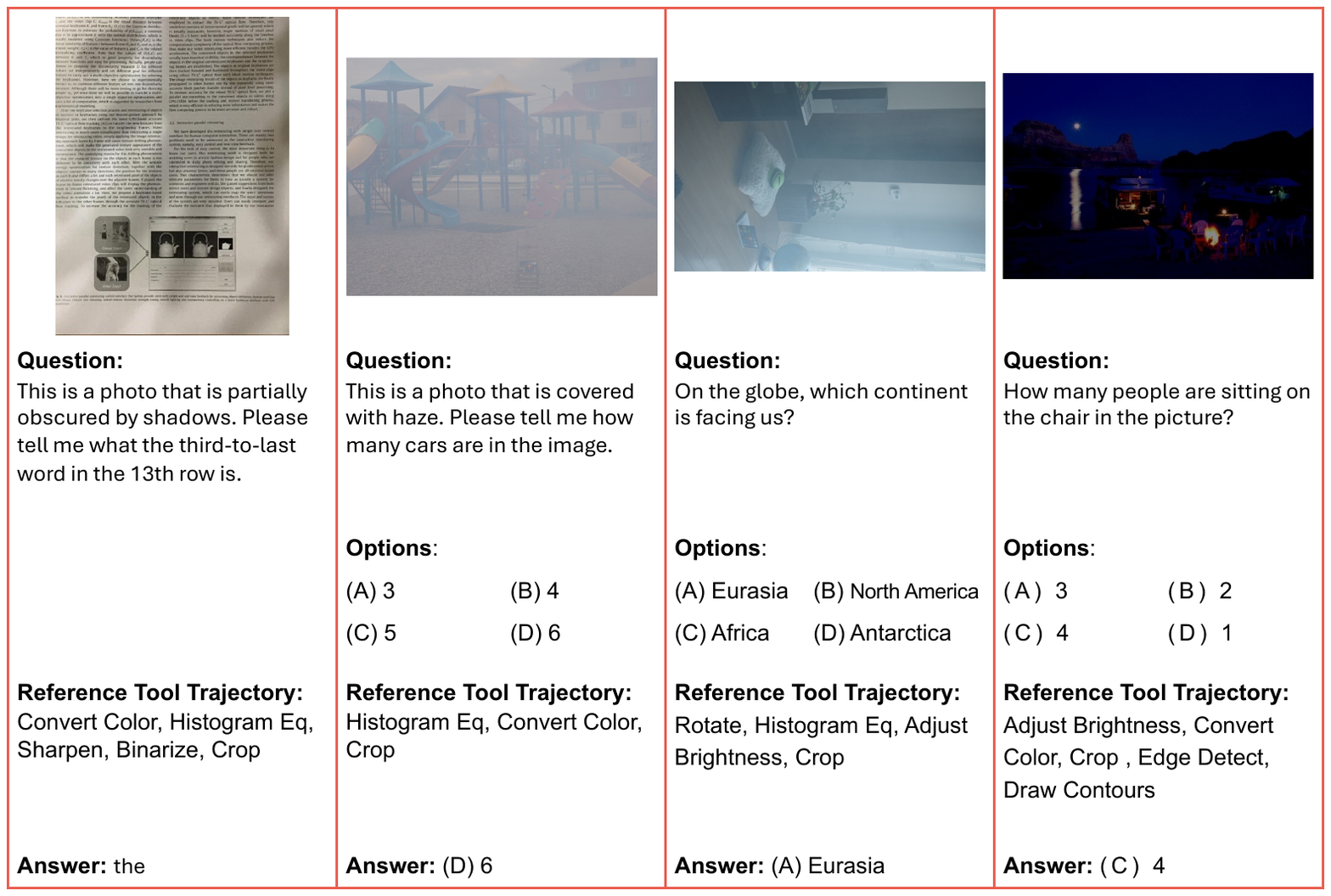}
    \caption{Supplementary examples of Perceptual Restoration tasks.}
    \label{fig:appendix_percept}
\end{figure}

\begin{figure}[t]
    \centering
    \includegraphics[width=0.95\linewidth]{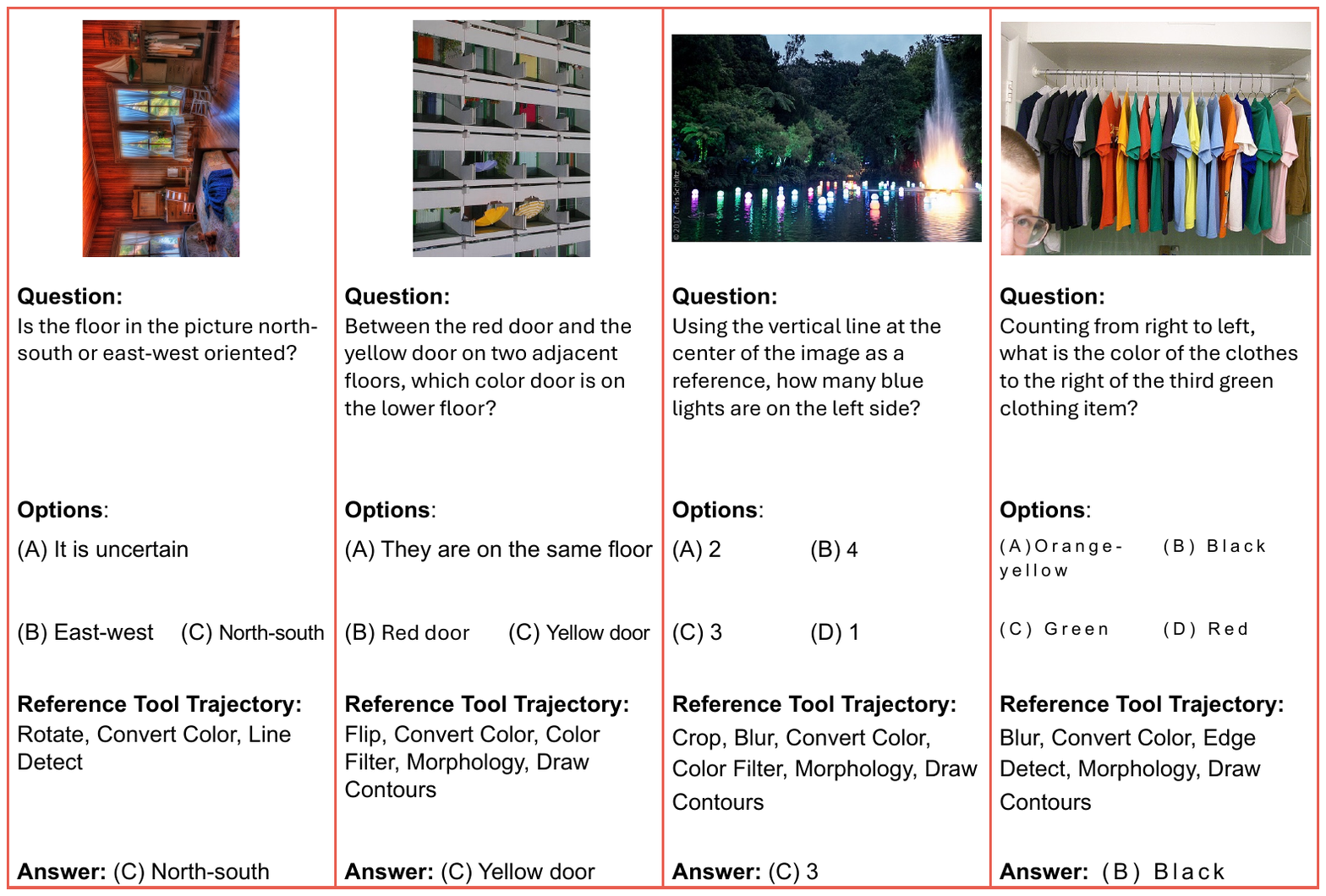}
    \caption{Supplementary examples of Spatial Reasoning tasks.}
    \label{fig:appendix_spatial}
\end{figure}

\clearpage
\section{Detailed Model Reasoning  Process}\label{app:detaled_model_response}
\begin{figure}
    \centering
    \includegraphics[width=0.9\linewidth]{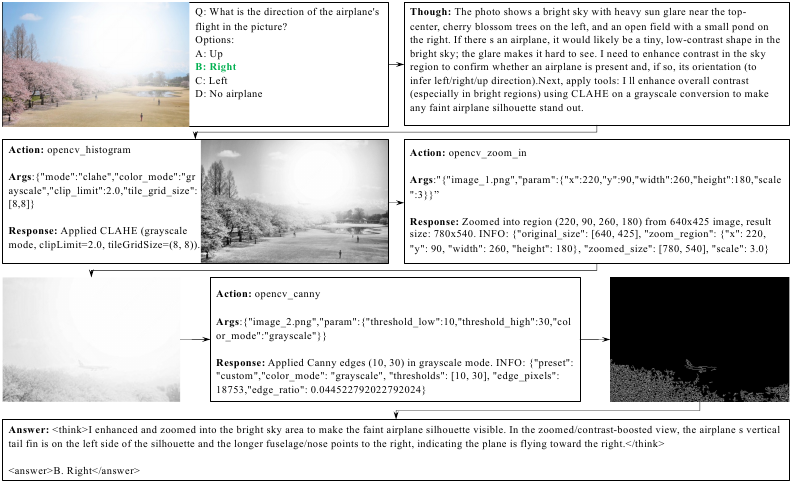}
    \caption{A case study of the reasoning process for Perceptual Restoration tasks.}
\end{figure}

\begin{figure}
    \centering
    \includegraphics[width=0.9\linewidth]{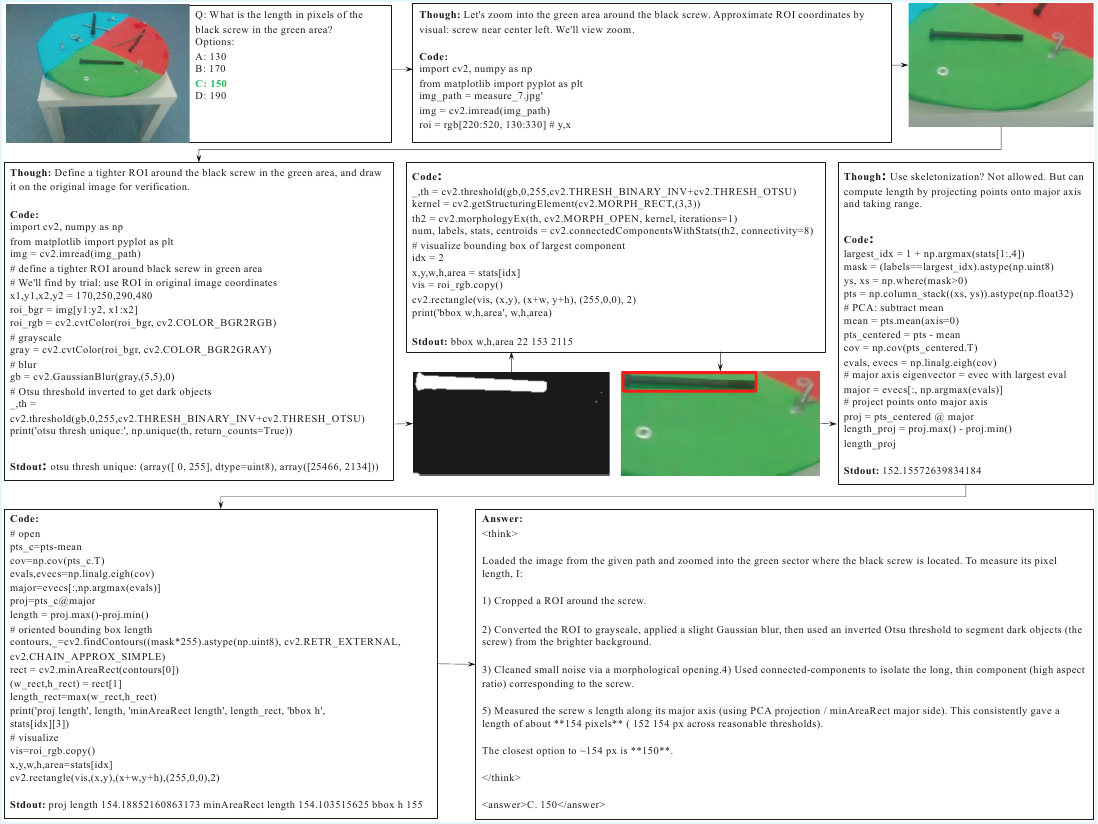}
    \caption{A case study of the reasoning process for Measure tasks.}
\end{figure}

\end{document}